\documentclass[letterpaper]{article} % DO NOT CHANGE THIS
\usepackage{aaai24}  % DO NOT CHANGE THIS
\usepackage{times}  % DO NOT CHANGE THIS
\usepackage{helvet}  % DO NOT CHANGE THIS
\usepackage{courier}  % DO NOT CHANGE THIS
\usepackage[hyphens]{url}  % DO NOT CHANGE THIS
\usepackage{graphicx} % DO NOT CHANGE THIS
\usepackage{natbib}  % DO NOT CHANGE THIS AND DO NOT ADD ANY OPTIONS TO IT
\usepackage{caption} % DO NOT CHANGE THIS AND DO NOT ADD ANY OPTIONS TO IT
\usepackage{algorithm}
\usepackage{algorithmic}
\usepackage{amsmath}
\usepackage{amssymb}

\usepackage{subcaption}
\usepackage{array}
\usepackage{makecell}
\usepackage{multirow}
\usepackage{comment}
\usepackage{bm}
\usepackage{xcolor}
\usepackage{mathtools}
\newcommand{\etal}{\textit{et al}. }
\newcommand{\ie}{\textit{i}.\textit{e}., }
\newcommand{\eg}{\textit{e}.\textit{g}., }
\newcolumntype{C}[1]{>{\centering\arraybackslash}m{#1}}
\newcolumntype{L}[1]{>{\arraybackslash}m{#1}}

\usepackage{newfloat}
\usepackage{listings}
\DeclareCaptionStyle{ruled}{labelfont=normalfont,labelsep=colon,strut=off} % DO NOT CHANGE THIS
\floatstyle{ruled}
\newfloat{listing}{tb}{lst}{}
\floatname{listing}{Listing}
\title{Spectrum Translation for Refinement of Image Generation (STIG)\\Based on Contrastive Learning and Spectral Filter Profile}
\author {
    Seokjun Lee\textsuperscript{\rm 1, \rm 2},
    Seung-Won Jung\textsuperscript{\rm 2},
    Hyunseok Seo\textsuperscript{\rm 1}\protect\thanks{Corresponding author.}
}
\affiliations {
    \textsuperscript{\rm 1}Biomedical Research Division, Korea Institute of Science and Technology\\
    \textsuperscript{\rm 2}School of Electrical Engineering, Korea University\\
    ykykyk112@kist.re.kr, swjung83@korea.ac.kr, seo@kist.kr
}

\usepackage{bibentry}
\begin{document}

\maketitle

\begin{abstract}
Currently, image generation and synthesis have remarkably progressed with generative models. Despite photo-realistic results, intrinsic discrepancies are still observed in the frequency domain. The spectral discrepancy appeared not only in generative adversarial networks but in diffusion models. In this study, we propose a framework to effectively mitigate the disparity in frequency domain of the generated images to improve generative performance of both GAN and diffusion models. This is realized by spectrum translation for the refinement of image generation (STIG) based on contrastive learning. We adopt theoretical logic of frequency components in various generative networks. The key idea, here, is to refine the spectrum of the generated image via the concept of image-to-image translation and contrastive learning in terms of digital signal processing. We evaluate our framework across eight fake image datasets and various cutting-edge models to demonstrate the effectiveness of STIG. Our framework outperforms other cutting-edges showing significant decreases in FID and log frequency distance of spectrum. We further emphasize that STIG improves image quality by decreasing the spectral anomaly. Additionally, validation results present that the frequency-based deepfake detector confuses more in the case where fake spectrums are manipulated by STIG.
\end{abstract}

\section{Introduction}

\begin{figure}[!t]
\centering
    \begin{tabular}{@{\hskip -0.1cm}L{0.0001cm}C{0.425\columnwidth}C{0.425\columnwidth}}
        \rotatebox{90}{CycleGAN} &
        \includegraphics[width=0.445\columnwidth]{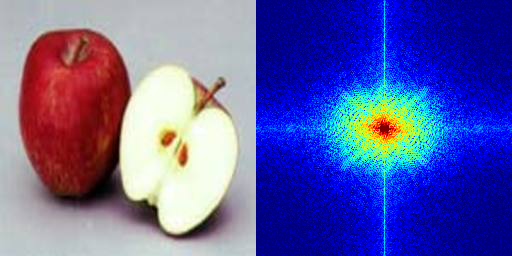} &
        \includegraphics[width=0.445\columnwidth]{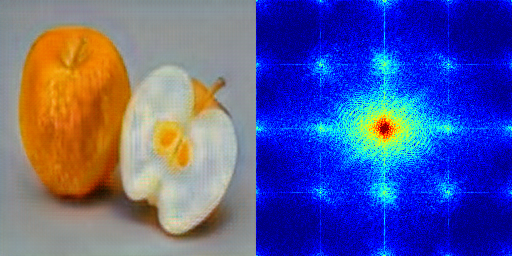} \\
        \rotatebox{90}{StarGAN} &
        \includegraphics[width=0.445\columnwidth]{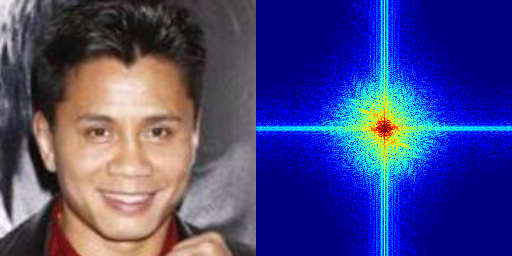} &
        \includegraphics[width=0.445\columnwidth]{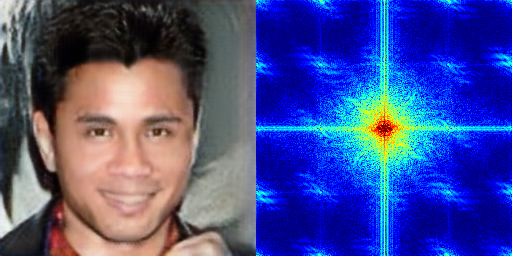} \\
        \rotatebox{90}{DDPM} &
        \includegraphics[width=0.445\columnwidth]{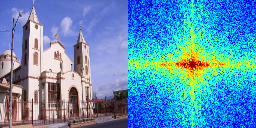} &
        \includegraphics[width=0.445\columnwidth]{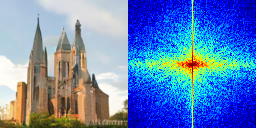} \\
        {} & \;Real (Image / Spectrum) & \;Fake (Image / Spectrum)
    \end{tabular}
    \caption{Spectral discrepancies between the real and generated image in various generative networks. CycleGAN \cite{cyclegan} and StarGAN \cite{stargan}, which include the transposed convolution layer (row 1 and row 2), produce the grid-based aliasing in the spectrum of the generated image. On the other hand, DDPM (row 3) \cite{ddpm} includes another type of discrepancy, the lack of high-frequency components.}
    \label{figure:example}
\end{figure}

In recent years, image generation and synthesis have noticeably progressed with various generative approaches \cite{dcgan, ddpm, score}. Further, many studies have shown ways to make the generative model produce more photo-realistic images which are not distinguishable by the human visual system \cite{stylegan, stylegan2, starganv2, glide}. In spite of their remarkable success, it is easy to find apparent discrepancies in the frequency spectrum of generated images. First, an upsampling operation in the most generative network can have aliasing on the spectrum (\ie checkerboard artifact in image domain) and insufficient high-frequency components \cite{autogan, leveraging}. In addition, the spectral disparity has been reported in the diffusion model, which currently stands out in image generation and synthesis fields. Researchers note that generated images from the diffusion model have insufficient high-frequency components \cite{sd}. Therefore, spectrum discrepancy has been an intrinsic challenge of the generative models. We visualized examples of these discrepancies in Fig.~\ref{figure:example}.

To close gaps in the frequency domain, several studies suggest approaches that modify the generative network architecture or objective function providing more guidelines for the real spectrum \cite{durall, disc, ffl}. Although they can restore the distortion in the power spectrum, there is still room for improvement in terms of frequency domain. To overcome these partial success of the conventional methods, we analyze the current generative networks in the frequency domain. We first check the spectrum profiles of the filters with respect to the kernel size to show the limitation of the conventional spatial domain approaches. We further revisit the denoising process of diffusion models using the Wiener filter-based approach so that the intrinsic problem of diffusion models (\ie insufficient high-frequencies) in the frequency domain comes out. Inspired by these analyses, we propose a frequency domain framework, \textbf{STIG}, that generally reduces the spectral distortion of the generated image. We evaluate the effectiveness of the proposed framework on various generative adversarial networks (GANs). Furthermore, we extend our experiments to another mainstream of image generation, diffusion models (DMs). The proposed framework successfully closes the gaps in the frequency domain while improving the quality of the generated image in the spatial domain as well.

The media forensic community has used spectral features to detect the synthetic image \cite{frepgan, morphing, marra}. Thus, spectral discrepancies can be clues for promising fake image detection not only for the GANs but also DMs \cite{leveraging, toward}. In this study, our framework is validated by FID, log frequency distance of spectrum, and frequency-based fake image detector for both GANs and DMs. In particular, the generated images manipulated by STIG make severe confusion in frequency-based fake image detection. Our contributions can be summarized as follows:
\begin{itemize}
  \item We analyze the intrinsic limitation of generative models based on the signal processing theorem to explain the spectral discrepancy of the current generative models.
  \item We propose a novel framework, \textbf{STIG}, that effectively reduces the intrinsic discrepancy of the generated images observed in the frequency domain.
  \item We verify the effectiveness of the proposed framework on various GANs and another mainstream of image generation, DMs.
  \item We present that between the real and the generated, STIG makes difficult discrimination of the frequency-based detectors.
\end{itemize}

\section{Related Work}
\subsection{Frequency Discrepancy in Generative Networks}
Earlier, Odena \etal \shortcite{odena} noted the checkerboard artifacts in the generated images. They found that the transposed convolution is relevant to these artifacts. Further studies \cite{leveraging, durall, autogan} showed that the artifacts originate from the high-frequency replica during the upsampling. Especially, Durall \etal \shortcite{durall} presented that the image generated with an interpolation does not include enough high-frequency components. Wang \etal \shortcite{wang} presented various types of spectral discrepancy in the GANs. Recently, researchers explore the spectrum inconsistency of the DMs. Rissanen \etal \shortcite{ihdm} analyzed the diffusion process in the frequency domain. They stated that the DMs have an inductive bias; during the reverse process, low-frequency components are synthesized first and high-frequencies are added to it later. Yang \etal \shortcite{sd} has also presented an experiment that the DMs have intrinsic defects in producing high-frequencies.

To improve the quality of the generated images, many studies have made an effort to narrow gaps in the frequency domain. Durall \etal \shortcite{durall} proposed a spectral regularization with the one-dimensional power distribution. Following studies \cite{disc, ssdgan} proposed a discriminator that enables the network to utilize the power spectral density (PSD). Jiang \etal \shortcite{ffl} defined a frequency-level objective function, focal frequency loss (FFL), to increase the flexibility in frequency components. Recently, the frequency domain approach to correct the spectral discrepancy has been reported \cite{thinktwice}. They refined the magnitude spectrum of the generated image using CycleGAN \cite{cyclegan} architecture. Although current SOTA methods improve the quality of image generation and reduce spectral discrepancies, most of them still have problems of aliasing-related artifacts and insufficient high-frequency components. Besides, they basically focused on GANs. So, we propose STIG compatible to both GANs and DMs.

\subsection{Frequency-Based Deepfake Detection} 
Due to the social impact of photo-realistic generated images in terms of security and privacy, computer vision and forensic research groups have paid attention to detecting the generated images \cite{cozzolino, spatial1, colorcue}. In addition to the earlier spatial domain approaches, many studies attempted to utilize spectral discrepancies. The periodic patterns on the spectrum and inconsistency of the spectral distribution can be a good clue for detecting fake images. Early studies boosted the detection performance by employing the magnitude of the spectrum as a detector input \cite{leveraging, autogan}. In particular, Frank \etal \shortcite{leveraging} pointed out the effectiveness of the frequency-based detector for performance and computational efficiency.
\begin{comment}
    In this paper, we review the credibility of the frequency domain detection approaches on various GANs and DMs, raising an alarm in our society.
\end{comment}

\section{Method}

\begin{figure}[t]
\centering
   \begin{subfigure}{1.0\linewidth}
   \includegraphics[width=1.0\linewidth]{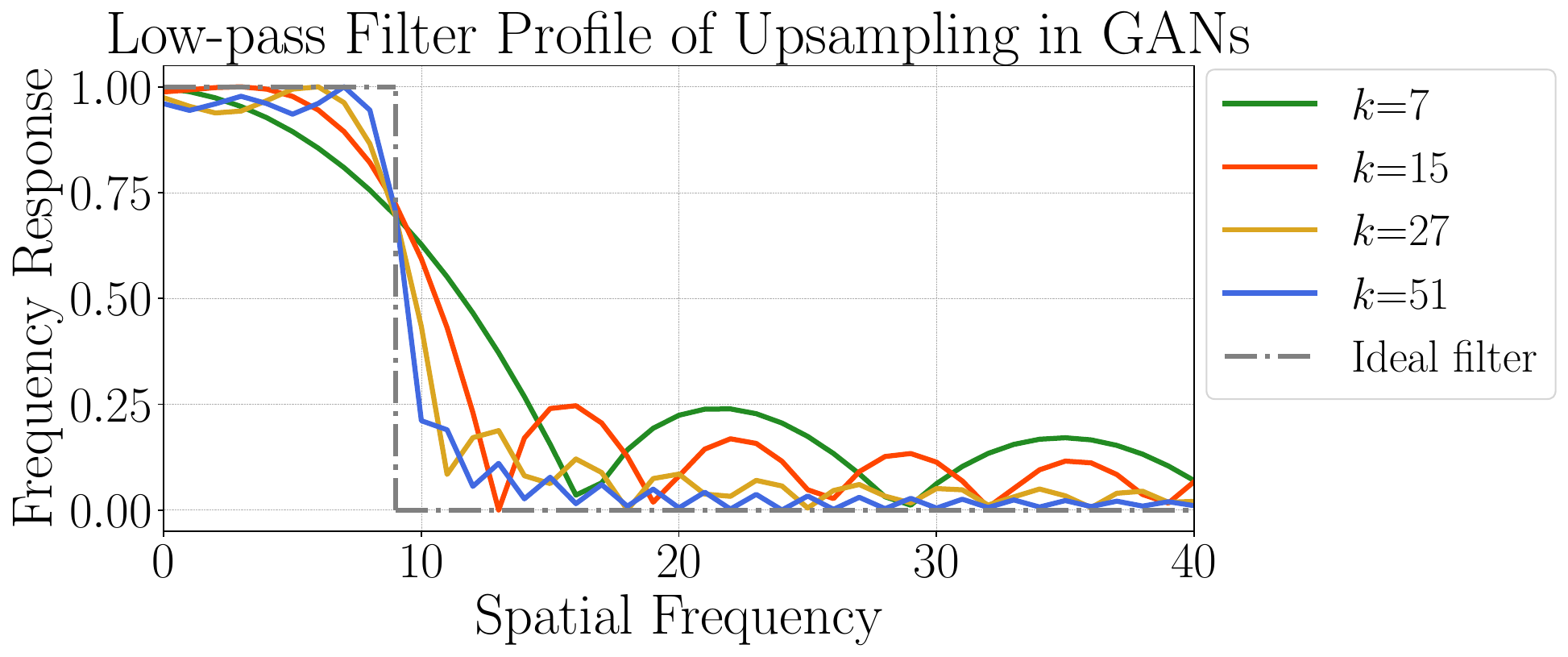}
   \caption{} \label{figure:lowpass_a}
   \end{subfigure}
   \begin{subfigure}{1.0\linewidth}
   \includegraphics[width=1.0\linewidth]{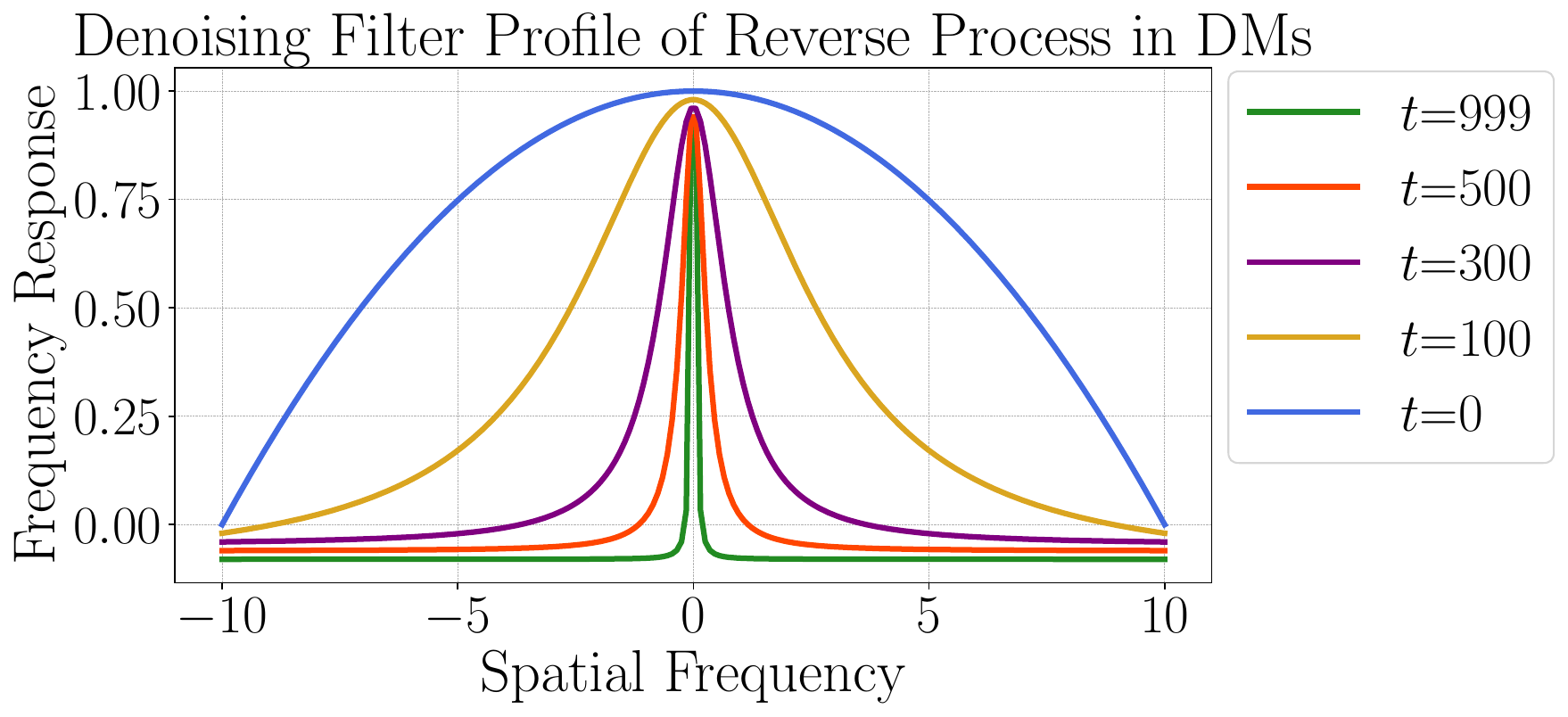}
   \caption{} \label{figure:lowpass_b}
   \end{subfigure}
\caption{(a) Estimation of an ideal low-pass filter by simulation with an example sinc function in the spatial domain. The sinc function with a finite kernel size causes an obvious ripple (\ie fluctuation pattern) in the cut-off frequency band. (b) Frequency response of the denoising filter for reverse process in diffusion models. The filter still blocks the high-frequency band even if at the end of the reverse process. We notice $t \in [0, 1000]$ in this example.}
\label{figure:lowpass}
\end{figure} 

\subsection{Frequency Domain Refinement}
\subsubsection{Frequency Analysis of GANs.}
Basically, ideal upsampling requires an ideal low-pass filter to remove the aliasing. Upsampling with classical interpolation (\eg bi-linear or bi-cubic) effectively suppresses aliasing, however, produces lack of high-frequency components. Although transposed convolution can be commonly adopted in GANs for image details, there is also a limitation that cannot be overlooked. In Fig.~\ref{figure:lowpass_a}, we demonstrate the effect of the kernel size on profile of an ideal low-pass filter from a sinc function. Although all kernels in Fig.~\ref{figure:lowpass_a} represent the low-pass filter, some kernels have severe stop-band ripples, \ie the fluctuation after the cut-off frequency. This observation supports the following limitation of the existing approaches. Conventional CNN kernels commonly used in deep neural networks are difficult to remove high-frequency replicas due to small kernel size. Previous studies for reducing spectral discrepancies mainly focus on modifying the network architecture and adopting regularization of the training objective in the spatial domain. Their goal can be described as encouraging the network to build the appropriate filters in the frequency domain, \eg ideal rectified filter. However, kernels with a finite kernel size cannot cut off the aliasing perfectly due to the incomplete shape of the low-pass filter in the frequency domain.

\subsubsection{Frequency Analysis of DMs.}
We further extend the frequency analysis to the diffusion models. We can show an intrinsic reason for spectrum discrepancy in the diffusion model by modification of the Wiener filter-based approach \cite{sd} for the denoising (reverse) process. First, if we assume the linear denoising Wiener filter $\mathbf{h}_{t}$, the objective function in the conventional diffusion model \cite{ddpm} can be rewritten as:
\begin{equation}
    \lVert \bm{\epsilon}
    -\gamma \cdot (\mathbf{1} - \mathbf{h}_{t}) \ast \mathbf{x}_{t} \rVert^{2}
    \label{equation:ddpm_loss}
\end{equation}
where $\bm{\epsilon}$ is the Gaussian noise, $\bm{\epsilon} \sim \mathcal{N}(\mathbf{0},\,\mathbf{I})$, and $\gamma = 1/\sqrt{1-\Bar{\alpha}_{t}}$. $\mathbf{X}_{t}$ and $\mathbf{H}_{t}$ represent the frequency response of $\mathbf{x}_{t}$ and $\mathbf{h}_{t}$ in diffusion models, respectively. Here, the Wiener filter denoises $\mathbf{x}_{t}$ by optimizing $\lVert\sqrt{\Bar{\alpha}_t}\mathbf{x}_{0}-\mathbf{h}_{t}\ast \mathbf{x}_{t} \rVert^{2}$, where $\sqrt{\Bar{\alpha}_t}$ is a scaling factor in original DDPM study \cite{ddpm}. From the power law, $\mathbf{X}_{0} \approx 1/f^{2}$ \cite{powerlaw}, we can define the conjugate transposed Wiener filter $\mathbf{H}_{t}^{\ast}$ as follows:
\begin{equation}
    \mathbf{H}_{t}^{\ast}(f) = \frac{\Bar{\alpha}_{t}} {{\Bar{\alpha}_{t}} + (1 - \Bar{\alpha}_{t}) \cdot f^{2}}
    \label{equation:wiener filter}
\end{equation}
More detailed expression is included in the supplementary materials. Since $\lvert \mathbf{H}_{t}^{\ast} \rvert = \lvert \mathbf{H}_{t} \rvert$, we can describe the magnitude profile of Wiener filter $\lvert \mathbf{H}_{t} \rvert$ in Fig.~\ref{figure:lowpass_b}. In DMs, the denoising filter focuses on composing low-frequency at the beginning of the sampling process and gradually adds high-frequencies as $t$ decreases. We can easily see this in common DMs. However, there is a wideband low-pass filter even if at the end of the reverse process, $t=0$. Therefore, generated images can suffer from insufficient high-frequency details. Based on the aforementioned frequency domain analyses, we precisely look up the spectrum in the frequency domain rather than in the spatial domain.

\subsection{STIG via Contrastive Learning}
\subsubsection{Generative Adversarial Learning.}
To reduce the spectral discrepancy in the generated images, we first assume the generated and real images to be bilateral distributions. One could recognize that the former has spectral anomalies and the latter is faultless. Then, we can define the spectrum domains $X \subset \mathbb{C}^{H\times W\times C}$ and $Y \subset \mathbb{C}^{H\times W\times C}$ which correspond to generated spectrum and real spectrum, respectively. From the domains, unpaired samples are given, $\mathbf{x} \in X$ and $\mathbf{y} \in Y$. Our goal is to find the mapping function between both domains to refine a generated spectrum $\mathbf{x}$. Therefore, we adopt an adversarial loss \cite{lsgan}, defined as:
\begin{equation}
    \mathcal{L}_{adv}(G,D,X,Y)= \mathbb{E}_{\mathbf{y}\sim Y}(D(\mathbf{y})-1)^{2}+ \mathbb{E}_{\mathbf{x}\sim X}(D(G(\mathbf{x})))^{2}
    \label{equation:6}
\end{equation}
where $G$ denotes the mapping function between $X$ and $Y$; $D$ aims to distinguish $G(\mathbf{x})$ from $\mathbf{y}$.

\begin{figure*}[!ht]
\begin{center}
   \includegraphics[width=1.0\linewidth]{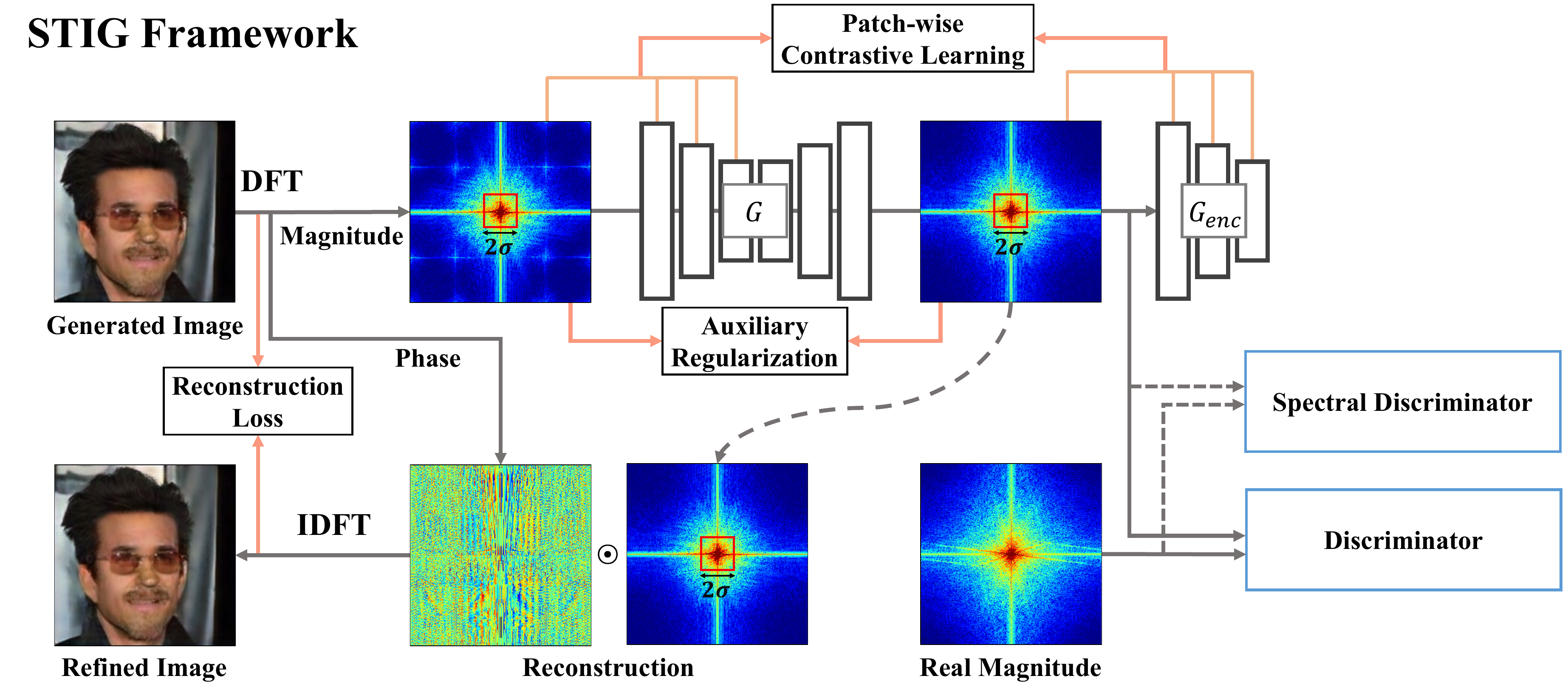}
\end{center}
\caption{Framework of our \textbf{STIG}. We exploit the magnitude spectrum as an input of our framework using the discrete Fourier Transform to reduce the spectral discrepancies. The spectrum of the generated image is translated into the domain of real spectrum. We, then, obtain the refined generated image by applying the inverse discrete Fourier Transform.}
\label{figure:framework}
\end{figure*}

\subsubsection{Patch-Wise Contrastive Learning.}
When the network decreases the discrepancy in the generated spectrum, it is important to preserve the original frequency components which are not related to the discrepancy. This is a significant issue in our study because a single frequency component affects the overall image contents. To resolve this challenge, we employ contrastive learning \cite{contrastive}. For each spectrum 
of images, we can define the frequency patch by extracting the spectral feature during the feedforward process. Since frequency patches in the same location share equivalent spatial frequencies, a patch that has abnormal frequencies (\eg aliased frequency and insufficient high-frequency) can be recognized as a negative sample in terms of contrastive learning. We define this contrastive tactic \cite{cut, dualcut} as follows:
\begin{equation}
    \mathcal{L}_{pcl}(G,H,X)= \mathbb{E}_{\mathbf{x}\sim X}\sum_{l=1}^{L}\sum_{s=1}^{S_{l}} \ell_{\text{CE}} (\hat{\mathbf{f}}_{l}^{s},\mathbf{f}_{l}^{s},\mathbf{f}_{l}^{S \setminus s})
    \label{equation:7}
\end{equation}
where $H$ denotes a two-layer MLP. When the input spectrum $\mathbf{x}$ passes through the $G_{\text{enc}}$ which implies the encoder part of $G$, frequency patches $\mathbf{f}_{l}=H(G_{\text{enc}}^{l}(\mathbf{x}))$ are extracted from $l$-th internal layer of $G_{\text{enc}}$. Especially, $\mathbf{f}_{l}^{s}$ represents $s$-th patch of $\mathbf{f}_{l}$ where $S \setminus s$ stands for the set $S$ excluding an element $s$ and $S_{l}$ denotes the pre-defined number of patches extracted from $l$-th layer. Similarly, we extract frequency patches of the translated output, $\hat{\mathbf{f}}_{l}=H(G_{\text{enc}}^{l}(G({\mathbf{x}})))$. The function, $\ell_{\text{CE}}(\cdot)$ in Eq.~\ref{equation:7} is a noise contrastive estimation (NCE) framework \cite{noisecontrastiveestimation} used for contrastive learning. By minimizing NCE framework, $G$ maximizes mutual information of the input-output pair $(\mathbf{f}_{l}^{s},\hat{\mathbf{f}}_{l}^{s})$; the frequency patch $\hat{\mathbf{f}}_{l}^{s}$ pulls a positive sample $\mathbf{f}_{l}^{s}$ while pushes negative samples $\mathbf{f}_{l}^{S \setminus s}$. Therefore, the mapping function $G$ is able to recognize the spectral discrepancies and finely keep the original frequency contents of an input spectrum while reducing the spectral anomaly.

\subsubsection{STIG Framework.}
We illustrate the framework of spectrum translation, STIG, in Fig.~\ref{figure:framework}. To shift the process domain, we first transform an input image into the frequency domain using the discrete Fourier Transform. STIG especially utilizes the magnitude spectrum because it includes most of the energy of spatial frequency and represents the spectral anomaly well. Then, the generator $G$ translates the spectrum of the generated image $\mathbf{x}$ into $Y$, the domain of real spectrum. In spectrum translation, patch-wise contrastive loss enables $G$ to strictly discriminate the abnormal frequencies in $\mathbf{x}$ and recover it, \eg erasing the bundle of aliasing or filling insufficient high-frequencies. Meanwhile, we adopt the spatial domain constraint between an input image and a refined image. We implement this using the SSIM reconstruction score \cite{ssim}. Therefore, the objective function for spectrum translation is defined as:
\begin{equation}
\mathcal{L}_{trans}=\mathcal{L}_{adv}+\mathcal{L}_{pcl}+\mathcal{L}_{rec}
\end{equation}

\subsubsection{Auxiliary Regularizations.}
Another key point to close the gap in the frequency domain is matching the distribution of the power spectrum. It is well known that the power spectral density of natural images follows a power law ${1/f^\alpha}$, where $\alpha$ = 2, approximately \cite{powerlaw}. To adjust the power spectral density of the generated spectrum, we introduce the chessboard integration which follows Chebyshev (\ie chessboard) distance \cite{chebyshev} and the spectral discriminator:
\begin{equation}
\begin{aligned}
    A_{k}(F)=\sum_{u=-k}^{k}\sum_{v=-k}^{k}\left\vert F(u,v) \right\vert, 
    \text{\,} k=0,\cdots,M/2-1.
\end{aligned}
\end{equation}
\begin{equation}
    \begin{aligned}
    CI_{k}(F)=A_{k}(F)-A_{k-1}(F),
    \text{\,} k=1,\cdots,M/2-1.
    \end{aligned}
\end{equation}
where $\left\vert F(u,v) \right\vert$ denotes the magnitude of the discrete Fourier Transform whose size is $M \times N$. Especially, $CI_{0}(F)$ is defined as $A_{0}(F)$ which implies the magnitude of DC frequency. More details are included in the supplementary materials. Since the aliased signal during the upsampling is lying on a Cartesian grid, our chessboard integration easily grabs the aliasing pattern in addition to the power disparity. We use the same adversarial objective in Eq.~\ref{equation:6} for training the spectral discriminator $D_{s}$, which is formulated as:
\begin{align}
    \mathcal{L}_{spec}(G,D_{s},X,Y)= \mathbb{E}_{\mathbf{y}\sim Y}(D_{s}(CI(\mathbf{y}))-1)^{2} \\ + \mathbb{E}_{\mathbf{x}\sim X}(D_{s}(CI(G(\mathbf{x}))))^{2} \nonumber
\end{align}
Here, $CI_{k}(\cdot) \in CI(\cdot)$.

\begin{table*}[!ht]
\begin{center}
    \begin{tabular}{C{2.9cm}| C{1.2cm} | C{1.15cm} C{1.15cm} C{1.15cm} C{1.15cm} C{1.15cm} C{1.15cm} C{1.15cm} C{1.15cm}}
    \Xhline{2\arrayrulewidth}
        \multirow{2}{*}{Method} & \multirow{2}{*}{Domain} & \multicolumn{2}{c}{CycleGAN} & \multicolumn{2}{c}{StarGAN} & \multicolumn{2}{c}{StarGAN2} & \multicolumn{2}{c}{StyleGAN} \\ 
        {} & {} & FID$\downarrow$ & LFD$\downarrow$ & FID$\downarrow$ & LFD$\downarrow$ & FID$\downarrow$ & LFD$\downarrow$ & FID$\downarrow$ & LFD$\downarrow$ \\ \hline\hline
        Original Generated & - & 58.75  & 15.31  & 192.48  & 15.43 &  15.03  & 16.10  & 3.59  & 13.65   \\ 
        Durall \etal & S & 64.53  & 16.11  & 74.00  & 14.71 &  21.03  & 16.00  & 3.46  & 13.47    \\ 
        Jung and Keuper & S & 33.83  & 16.60  & 120.87  & 15.28 &  19.01  & 15.99  & 14.12  & 14.56   \\ 
        FFL & S & 50.29  & 15.62  & 184.84  & 15.31 &  17.74  & 16.08  & 3.98  & 13.76   \\
        SpectralGAN & F & 14.22  & 14.50  & 19.96  & 15.02 &  7.31  & 14.44  & 3.91  & 12.38    \\ 
        STIG (ours) & F & \textbf{6.97}   & \textbf{12.77}  & \textbf{6.30}  & \textbf{12.87} &  \textbf{3.10}  & \textbf{12.58}  & \textbf{2.37}  & \textbf{12.26}  \\ \Xhline{2\arrayrulewidth}
    \end{tabular}
\caption{FID and log frequency distance (LFD) of the magnitude spectrum for GANs. The upper three methods stand for the spatial domain method (S), \ie Durall \etal, Jung and Keuper, and FFL. On the other hand, SpectralGAN and STIG stand for the frequency domain method (F). The arrow implies the direction of the best score in FID and LFD, respectively.}
\label{table:gan_frequency}
\end{center}
\end{table*}

\begin{table*}[!ht]
\begin{center}
    \begin{tabular}{C{2.9cm}| C{1.15cm} C{1.15cm} C{1.15cm} C{1.15cm} C{1.15cm} C{1.15cm} C{1.15cm} C{1.15cm}}
    \Xhline{2\arrayrulewidth}
        \multirow{2}{*}{Method} & \multicolumn{2}{c}{DDPM-Face} & \multicolumn{2}{c}{DDPM-Church} & \multicolumn{2}{c}{DDIM-Face} & \multicolumn{2}{c}{DDIM-Church} \\ 
        {} & FID$\downarrow$ & LFD$\downarrow$ & FID$\downarrow$ & LFD$\downarrow$ & FID$\downarrow$ & LFD$\downarrow$ & FID$\downarrow$ & LFD$\downarrow$ \\ \hline\hline
        Original Generated & 43.54 & 15.28 & 59.66 & 15.49 & 51.49 & 15.23 & 55.25 & 15.48   \\ 
        STIG (ours)        & \textbf{21.91} & \textbf{14.52} & \textbf{22.53} & \textbf{15.38} & \textbf{8.56}  & \textbf{12.70} & \textbf{16.47} & \textbf{14.72}   \\ \Xhline{2\arrayrulewidth}
    \end{tabular}
\caption{FID and log frequency distance (LFD) of the magnitude spectrum for DMs.}
\label{table:ddpm_frequency}
\end{center}
\end{table*}

Basically, the low-frequency part is significant for spectrum translation and reconstruction into the image. Therefore, we exploit the regularization term that constrains the low-frequency distance to maintain the energy level of an input image as follows:
\begin{equation}
    \mathcal{L}_{lf}=\sum_{u=-\sigma}^{\sigma}\sum_{v=-\sigma}^{\sigma} (\left\vert F(u,v)\right\vert-\left\vert F_{r}(u,v)\right\vert)^{2}
\end{equation}
where $\left\vert F(u,v)\right\vert$ and $\left\vert F_{r}(u,v)\right\vert$ are an input magnitude and a refined magnitude spectrum, respectively; $\sigma$ denotes the bandwidth taken to estimate distance. To prevent including artifact components, the bandwidth $\sigma$ can be configured carefully with the number of upsampling layers in the source network. We typically select $\sigma=8$ for all experiments in this paper. Finally, the training objective for STIG is defined as: \begin{equation}
    \mathcal{L}_{total}=\mathcal{L}_{trans}+\mathcal{L}_{spec}+\mathcal{L}_{lf}
\end{equation}

\section{Experiments}
\subsection{Experimental Setup}
\subsubsection{Training Details.} We evaluate STIG on datasets generated from diverse generative networks, not only GANs but DMs as well. For GANs, we select models of CycleGAN, StarGAN, StarGAN2, and StyleGAN \cite{cyclegan, stargan, starganv2, stylegan} following a setting in \cite{thinktwice}. We further generate fake images from two popular diffusion models, DDPM and DDIM \cite{ddpm, ddim}. For training the DMs, we especially employ two benchmarks (\ie FFHQ \cite{stylegan} and LSUN Church \cite{lsun}) to investigate the generalization ability of the proposed framework on diffusion models. We set the resolution to $128\times128$ on DMs and $256\times256$ on GANs. STIG is composed of the generator $G$ and two kinds of discriminators, $D$ and $D_{s}$ for spectrum translation. We apply a Nested U-Net \cite{nestedunet} architecture to the generator and PatchGAN to the discriminator \cite{patchgan}. For the spectral discriminator, we employ a simple fully connected layer with the sigmoid activation. More details about implementation are included in the supplementary materials.

\subsubsection{Comparison Methods.}
We compare our STIG with SOTA methods which aim to correct the spectral discrepancy for generative adversarial networks. We first verify our method with spectral regularization methods \cite{durall, disc}. Then, the distance-based approach, focal frequency loss (FFL) \cite{ffl} is applied for another validation. Besides, a frequency domain approach, SpectralGAN \cite{thinktwice} is compared as well. For the diffusion models, we verify superiority of our STIG in comparison to the original generated images.

\subsubsection{Evaluation Metrics.}
In this paper, we assess STIG with respect to the spectral discrepancy, image quality, and performance of deepfake detectors. For evaluation of STIG in the frequency domain, we first employ FID \cite{fid} to calculate the similarity between the real spectrum and each fake spectrum. Then, we report log frequency distance (LFD) \cite{ffl} by assessing the logarithm difference in averaged spectra. We convert the magnitude spectrum to be log-scaled and normalize it to [-1, 1]. In the spatial domain, we also verify the refined image from STIG using FID to show the improvement of the proposed method. Finally, we evaluate it with the deepfake detectors, which can show the effect of STIG as well.

\begin{figure*}[!t]
\centering
    \begin{tabular}{@{\hskip-0.003cm} C{0.200\columnwidth}C{0.200\columnwidth}C{0.200\columnwidth}C{0.200\columnwidth} C{0.0001\columnwidth}C{0.200\columnwidth}C{0.200\columnwidth}C{0.200\columnwidth}C{0.200\columnwidth}}
        \includegraphics[width=0.230\columnwidth]{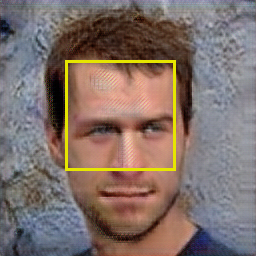} &
        \includegraphics[width=0.230\columnwidth]{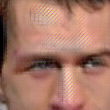} &
        \includegraphics[width=0.230\columnwidth]{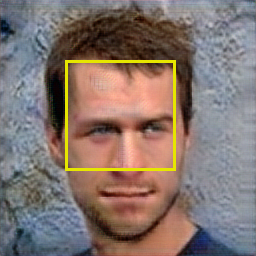} &
        \includegraphics[width=0.230\columnwidth]{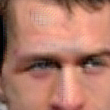} & {} &
        \includegraphics[width=0.230\columnwidth]{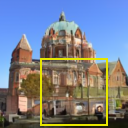} &
        \includegraphics[width=0.230\columnwidth]{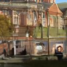} &
        \includegraphics[width=0.230\columnwidth]{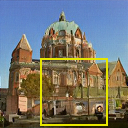} &
        \includegraphics[width=0.230\columnwidth]{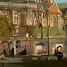}\\
        \includegraphics[width=0.230\columnwidth]{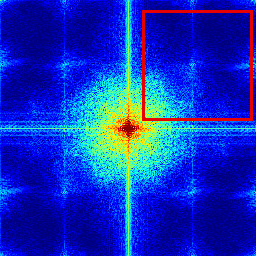} &
        \includegraphics[width=0.230\columnwidth]{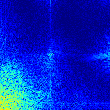} &
        \includegraphics[width=0.230\columnwidth]{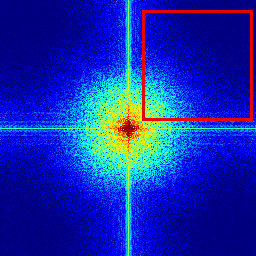} &
        \includegraphics[width=0.230\columnwidth]{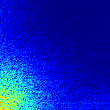} & {} &
        \includegraphics[width=0.230\columnwidth]{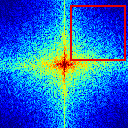} &
        \includegraphics[width=0.230\columnwidth]{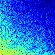} &
        \includegraphics[width=0.230\columnwidth]{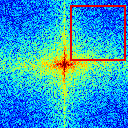} &
        \includegraphics[width=0.230\columnwidth]{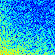} \\
        \multicolumn{2}{c}{Original (StarGAN)} & \multicolumn{2}{c}{STIG (StarGAN)} & {} & \multicolumn{2}{c}{\;\;Original (DDIM-Church)} & \multicolumn{2}{c}{\;STIG (DDIM-Church)}
    \end{tabular}
    \caption{STIG examples on StarGAN and DDIM-Church. We magnified the image and corresponding magnitude spectrum relevant to the spectral discrepancies (yellow and red boxed). The lefts indicate the original generated image and the corresponding spectrum. On the other hand, the rights indicate STIG-refined ones.}
    \label{figure:stig_example}
\end{figure*}

\subsection{Quantitative Analysis}
\subsubsection{Frequency Domain Results.}
We evaluate STIG in comparison with other methods that aim to reduce the spectral discrepancy in generated images. We report the FID between the real and fake magnitude spectrum and the LFD of the averaged spectrum in Table~\ref{table:gan_frequency}. It shows that the originally generated spectrum has a severe statistical inconsistency except for StyleGAN. As can be seen, the spatial domain methods mitigate spectral disparity, however, they rather deepen the discrepancy in some cases, \eg StarGAN2 and StyleGAN. On the other hand, the frequency domain methods successfully reduce the discrepancy in the frequency domain. Although SpectralGAN works well for most benchmarks, it makes worse in StyleGAN, where the disparity is relatively gentle. On the other hand, we show that our method effectively closes the gaps in the frequency domain, outperforming other methods. STIG presents well-generalized performance regardless of the architecture of the generative network. We also provide the frequency domain effect of STIG on diffusion models. Table~\ref{table:ddpm_frequency} shows that our method reduces the spectral discrepancy on diffusion models as well even though the source of the disparity is different from the GAN-based models. STIG decreases 67\% of FID and 1.04 of LFD on average for diffusion models.

\begin{table}[!t]
\begin{center}
    \begin{tabular}{C{1.1cm}|C{2.5cm}|C{1.5cm}C{1.5cm}}
    \Xhline{2\arrayrulewidth}
         \multirow{2}{*}{Type} & \multirow{2}{*}{Benchmark} & \multicolumn{2}{c}{FID of Image$\downarrow$} \\
        {} & {} & w/o STIG & w/ STIG \\ 
        \hline\hline
        \multirow{4}{*}{GANs} & CycleGAN  & 61.69 & \textbf{59.85} \\
        {} & StarGAN   & 26.67 & \textbf{25.38} \\ 
        {} & StarGAN2  & 13.52 & \textbf{13.36} \\ 
        {} & StyleGAN  & 20.17 & \textbf{19.92} \\ \hline
        \multirow{4}{*}{DMs} & DDPM-Face & 40.19 & \textbf{26.57} \\
        {} & DDPM-Church & 31.59 & \textbf{29.35} \\ 
        {} & DDIM-Face   & 36.70 & \textbf{22.19} \\ 
        {} & DDIM-Church & 27.62 & \textbf{22.53} \\ \Xhline{2\arrayrulewidth}
    \end{tabular}
    \caption{FID of the generated and STIG-refined image for various generative networks.}
\label{table:fid_image}
\end{center}
\end{table}

\subsubsection{Improvement on Image Quality.} From the duality of the discrete Fourier Transform, spectrum translation improves the quality of the generated image as well. The discrepancy in the frequency domain can be recognized in specific ways in the spatial domain, \eg periodic checkerboard artifact patterns or blurry texture. To evaluate the effect of STIG in the image domain, we calculate the FID from the originally generated images and STIG-refined images with real ones. Table~\ref{table:fid_image} presents the improvements of the generated images, processed with STIG in the frequency domain. STIG enhances the generated images for all benchmarks with diffusion models as well as generative adversarial networks. We emphasize that the proposed framework is especially effective for diffusion models. The diffusion-generated images are improved by 26.1\% in terms of FID on average.

\begin{table}[!t]
\begin{center}
    \begin{tabular}{C{2.2cm}|C{0.95cm}C{0.95cm}C{0.95cm}C{0.95cm}}
    \Xhline{2\arrayrulewidth}
        Loss function & \multicolumn{4}{c}{FID of Image$\downarrow$} \\
        {$\mathcal{L}_{lf}/\mathcal{L}_{spec}$} & O/O & O/X & X/O & X/X \\ 
        \hline\hline
        CycleGAN  & \textbf{59.85} & 60.62 & 60.70 & 60.62 \\
        StarGAN2  & \textbf{13.36} & 13.41 & 13.58 & 13.59 \\ 
        DDPM-Church & \textbf{29.35} & 29.42 & 31.15 & 31.25 \\ 
        DDIM-Face   & \textbf{22.19} & 22.23 & 25.02 & 24.90 \\ 
        \Xhline{2\arrayrulewidth}
    \end{tabular}
    \caption{Ablation study for the auxiliary regularization. We compare the performance of STIG under different combinations of auxiliary regularization terms on various
benchmarks.}
\label{table:ablation}
\end{center}
\end{table}

\begin{table*}[!t]
\begin{center}
\small
    \begin{tabular}{C{1.75cm}| C{1.3cm} C{1.3cm} C{1.3cm} C{1.3cm} C{1.4cm} C{1.4cm} C{1.3cm} C{1.35cm} | C{1.2cm}}
    \Xhline{2\arrayrulewidth}
        Method & CycleGAN & StarGAN & StarGAN2 & StyleGAN & DDPM-F* & DDPM-C* & DDIM-F* & DDIM-C* & Avg \\ \hline\hline
        Original Fake & 98.85\% & 100.00\% & 99.78\% & 97.24\% & 99.73\% & 99.27\% & 99.75\% & 99.39\% & 99.25\%  \\ 
        STIG        & 52.98\% & 50.00\% & 50.14\% & 53.78\% &  86.22\% & 62.03\% & 49.87\% & 53.52\% & 57.32\%  \\ \Xhline{2\arrayrulewidth}
    \end{tabular}
\caption{Detection accuracy of the CNN-based fake image detector in the frequency domain. F* and C* after the name of the diffusion models indicate Face and Church, respectively.}
\label{table:detection_cnn}
\end{center}
\end{table*}

\begin{table*}[!t]
\begin{center}
\small
    \begin{tabular}{C{1.75cm}| C{1.3cm} C{1.3cm} C{1.3cm} C{1.3cm} C{1.4cm} C{1.4cm} C{1.3cm} C{1.35cm} | C{1.2cm}}
    \Xhline{2\arrayrulewidth}
        Method & CycleGAN & StarGAN & StarGAN2 & StyleGAN & DDPM-F* & DDPM-C* & DDIM-F* & DDIM-C* & Avg \\ \hline\hline
        Original Fake & 100.00\% & 100.00\% & 99.98\% & 99.89\% & 99.97\% & 99.92\% & 99.97\% & 99.95\% & 99.96\%  \\ 
        STIG        & 50.25\% & 50.00\% & 50.16\% & 53.01\% &  70.55\% & 57.63\% & 49.96\% & 50.46\% & 54.00\%  \\ \Xhline{2\arrayrulewidth}
    \end{tabular}
\caption{Detection accuracy of the ViT-based fake image detector in the frequency domain.}
\label{table:detection_vit}
\end{center}
\end{table*}

In addition to the quantitative results, we provide the visual effect of STIG by displaying the qualitative outcomes. First, we present a STIG example on the generative adversarial network, StarGAN, on the left side of Fig.~\ref{figure:stig_example}. There are noticeable checkerboard artifacts (\ie checkerboard patterns on the human face) in the original generated image which correspond to grid and dot patterns in the spectrum. By adopting STIG, artifacts in the generated image are successfully reduced. The spectral anomalies (\eg grid and dots) in the spectrum are completely removed as well. Figure~\ref{figure:stig_example} also shows the visual example of the diffusion model on the right side. The original generated image from the DDIM-Church benchmark has insufficient high-frequency content. We can see that STIG obviously produces appropriate high-frequency components while maintaining original spatial frequencies. As a result, the reconstructed image from STIG has sophisticated image features (\eg edges and texture) like a real image. We include more visual examples in the supplementary materials.

\subsection{Ablation Study}
We conducted ablation experiments to analyze the effect of auxiliary loss terms of STIG. Because auxiliary terms are adopted to encourage the proposed framework to preserve fundamental frequencies during spectral translation, we evaluate them in the spatial domain using FID. In Table~\ref{table:ablation}, we present the performance of STIG under different combinations of auxiliary regularizations. For most benchmarks, the low-frequency loss $\mathcal{L}_{lf}$ improves the visual quality of the refined image. Although the spectral discriminator does not affect the quality by itself, it performs well combined with $\mathcal{L}_{lf}$. Especially, auxiliary terms are effective for diffusion models, \ie 8.2\% decrease of FID on average. 

\begin{figure}[!t]
\centering
    \begin{tabular}{@{\hskip-0.05cm} C{0.285\columnwidth}C{0.285\columnwidth}C{0.285\columnwidth}}
    \includegraphics[width=0.320\columnwidth]{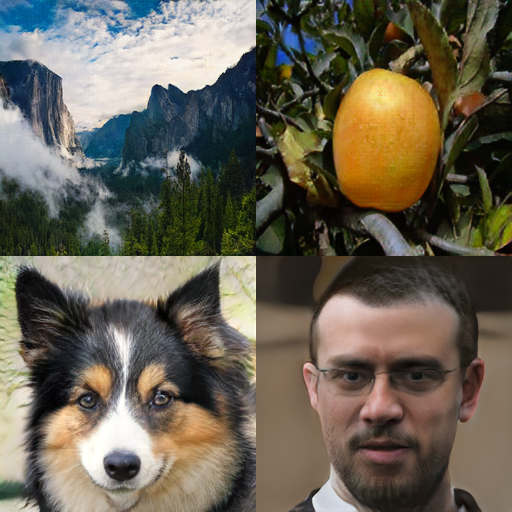} &
    \includegraphics[width=0.320\columnwidth]{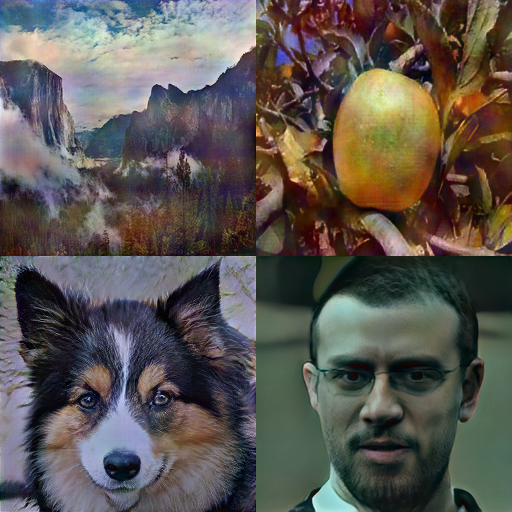} &
    \includegraphics[width=0.320\columnwidth]{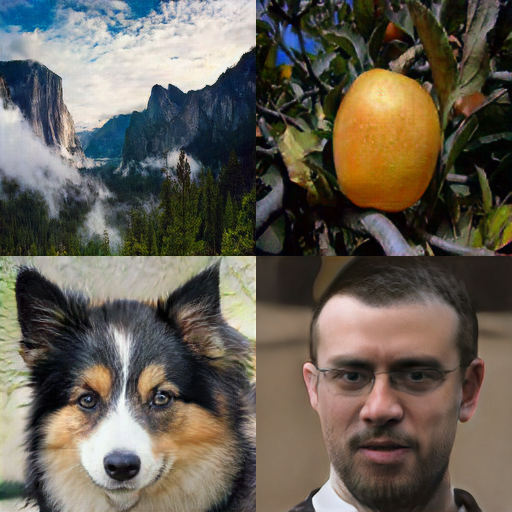} \\
    {\;\;\;\;Original Fake} & {\;\;\;SpectralGAN} & {\;\;\;STIG}
    \end{tabular}
    \caption{Comparison with a frequency domain method, SpectralGAN \cite{thinktwice}, for color tone. Examples are sampled from CycleGAN, StarGAN2, and StyleGAN benchmarks.}
    \label{figure:color_tone}
\end{figure}

Additionally, we compare STIG with another frequency domain approach, SpectralGAN \cite{thinktwice} in Fig.~\ref{figure:color_tone}. SpectralGAN breaks the color tone of original images. On the other hand, STIG preserves the energy level of color distribution by adopting auxiliary terms. $\mathcal{L}_{lf}$ aims to preserve the low-frequency energy of the original generated image while $\mathcal{L}_{spec}$ is matching the power distribution of the spectrum, \eg producing the insufficient high-frequencies. Thus, the two auxiliary terms enable more efficient learning.

\subsection{STIG on Frequency-Based Detector}
In this section, we investigate the effect of STIG on the frequency-based fake image detectors. First, we examine the conventional CNN-based frequency domain detector which is composed of shallow convolutional layers \cite{leveraging}. And then, we also consider the cutting-edge detector which has more network capacity and performs well in various computer vision tasks. We employ the ViT-B16 \cite{vit} for fake image detection as well. We trained and evaluated classifiers with the dataset consisting of real and fake (generated) image pairs. We provide more details for training detectors in the supplementary materials.

\subsubsection{Shallow CNN-Based Detector.} We provide the detection accuracy of the CNN-based detector in Table~\ref{table:detection_cnn}. The CNN-based detector well discriminates the fake spectrum from real spectrum in the frequency domain even though the detector has a very shallow architecture. The network can make the decision boundary by detecting spectral clues which are only in the spectrum of the fake image. However, the detector can't recognize the refined spectrum from real spectrum except DDPM-Face and DDPM-Church benchmark. The performance drop in average accuracy (\ie 41.93\%) implies that the CNN-based detector can be easily confused by STIG.

\subsubsection{ViT-Based Detector.} Table~\ref{table:detection_vit} shows the detection performance of the ViT-based frequency domain detector. We can see that the ViT-based detector perfectly detects the fake spectrum with an average accuracy of 99.96\%. After STIG applies spectrum translation to fake images, the average detection accuracy drops to 54\%. STIG makes a fool of the detector even though each detector is trained with generated images that are from diverse generative networks.
\subsubsection{Neutralized Frequency-Based Detector.} By investigating the effect of STIG on frequency-based detectors, we can show the promising detectors easily make confusion. First, CNN-based and ViT-based detector report perfect detection ability, around 99\%, for all benchmarks with the original generated. However, detection performance severely drops when improvement of the spectrum is applied.

\section{Conclusion}
In this paper, we propose the frequency domain approach that reduces the spectral discrepancy in the generated image. We generally analyze the intrinsic limitation of generative models in the frequency domain. Our method considerably improves spectral realism and image quality by directly manipulating the frequency components of the generated image. Experimental results on eight fake image benchmarks show that STIG significantly mitigates spectral anomalies. Our method also reports improved image qualities not only in generative adversarial networks but also in diffusion models. From the results with STIG, current frequency-based detectors are not enough to replace image-based detectors.

\section{Acknowledgments}
This work was partially supported by KIST Institutional Programs (2V09831, 2E32341, 2E32271) and by the Korea Medical Device Development Fund grant funded by the Korea government (202011A02).

\bibliography{aaai24}

\section{Implementation Details}
\subsection{Chessboard Integration}
\begin{figure}[!ht]
  \begin{center}
  \addtolength{\tabcolsep}{-4.5pt}
  \begin{tabular}{C{3.1cm}C{0.02cm}C{5.3cm}}
      \includegraphics[width=0.95\linewidth]{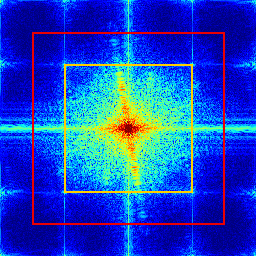}
    &
    {}
    &
      \includegraphics[width=0.95\linewidth]{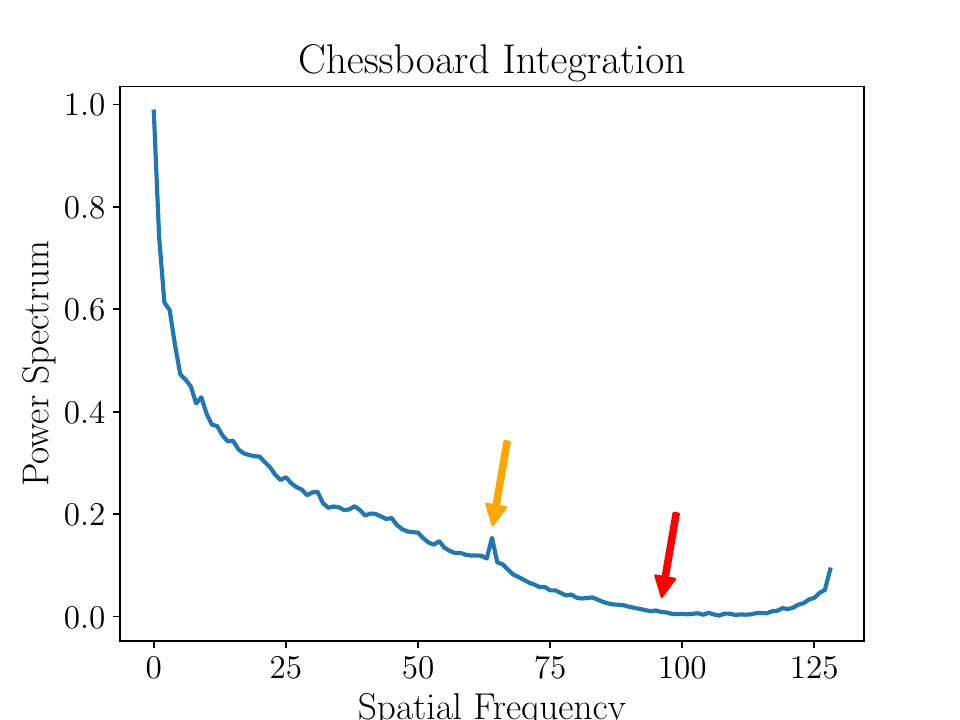}
  \end{tabular}
  \end{center}
  \caption{Example of the chessboard integration.}
  \label{figure:ci}
\end{figure}

We provide more details for the spectral discriminator. In the main document, we defined the chessboard integration, which follows Chebyshev (\ie chessboard) distance, as:
\begin{equation}
    A_{k}=\sum_{u=-k}^{k}\sum_{v=-k}^{k}\left\vert F(u,v) \right\vert, 
    \text{\;for\; } k=0,\cdots,M/2-1.
\end{equation}
\begin{equation}
    \begin{aligned}
    CI_{k}(A)=A_{k}-A_{k-1}, 
    \text{\;for\; } k=1,\cdots,M/2-1.
    \end{aligned}
\end{equation}
where $\left\vert F(u,v) \right\vert$ denotes the magnitude spectrum of the discrete Fourier Transform. Therefore, chessboard integration indicates the amount of each frequency energy accumulated along the grid-based rectangular trajectory in the spectrum. Besides, chessboard integration also easily grabs the aliased pattern since the aliased signal from upsampling appears on a Cartesian grid. In Fig.~\ref{figure:ci}, we depict an example of the chessboard integration for the magnitude spectrum. Chessboard integration results along the yellow and red lines in the spectrum (left) correspond to the same color arrows (right) in Fig.~\ref{figure:ci}. So, the aliasing in the 2D spectrum can easily turn into the peak in the 1D power spectrum.

\subsection{Training Details}
In this section, we provide additional implementation details of our study. In the main document, we described the objective function of STIG as follows:
\begin{equation}
\mathcal{L}_{trans}=\lambda_{1}\cdot\mathcal{L}_{adv}+\lambda_{2}\cdot\mathcal{L}_{pcl}+\lambda_{3}\cdot\mathcal{L}_{rec}
\label{equation:sup1}
\end{equation}
\begin{equation}
\mathcal{L}_{total}=\mathcal{L}_{trans}+\lambda_{4}\cdot\mathcal{L}_{spec}+\lambda_{5}\cdot\mathcal{L}_{lf}
\label{equation:sup2}
\end{equation}
where $\lambda_{n}$ denotes the weight of each loss term. In Eq.~\ref{equation:sup1}, we set $\lambda_{1}, \lambda_{3}, \lambda_{4}, \lambda_{5}=3$, and $\lambda_{2}=10$. For optimizing STIG, we employ Adam optimizer \cite{adam} with an initial learning rate of 8e-5. The optimizer keeps an initial learning rate during the first 20\% of the total training steps. After that, the learning rate linearly decreased to 0. The mini-batch size of the training samples is 1. We especially apply random rotation augmentation to the LSUN Church \cite{lsun} benchmark.

On the other hand, details of the fake image detectors are provided as follows. To train the frequency-based detectors, we adopt Adam \cite{adam} optimizer with an initial learning rate of 2e-4. We set the total training epoch to 20 and the mini-batch size to 32. The learning rate is decreased exponentially during the training. When we trained the fake image detectors, we used 80\% of the dataset for training and 20\% for evaluation. We implemented all experiments in our paper with the PyTorch \cite{pytorch} framework using an Nvidia RTX 3090ti.

\section{Derivation of Wiener Filter-Based Model}
In this section, we provide a detailed expression of the Wiener filter-based model in the main document.
\subsection{Review of Diffusion Model}
First, for the monotonically increasing value $\beta_{t}$, we can define the forward process in the diffusion model \cite{ddpm} as follows:
\begin{align}
    q(\mathbf{x}_{t}|\mathbf{x}_{0}) &= \mathcal{N}(\mathbf{x}_{t};\sqrt{\Bar{\alpha}_{t}}\mathbf{x}_{0},(1-\Bar{\alpha}_{t})\mathbf{I})\\
    \mathbf{x}_{t} &= \sqrt{\Bar{\alpha}_{t}}\mathbf{x}_{0}+\sqrt{(1-\Bar{\alpha}_{t})}\bm{\epsilon}
    \label{equation:forward_process}
\end{align}
where $\alpha_{t} \coloneqq 1-\beta_{t}$ and $\Bar{\alpha}_{t} \coloneqq \prod_{s=1}^{t} \alpha_{s}$; $\mathbf{x}_{0} \sim p(\mathbf{x}_{0})$ is an image sampled from the data distribution. Since the forward and reverse process follow Markovian, we can express $\mathbf{x}_{t}$ at an arbitrary timestep $t$ as a weighted sum of an image $\mathbf{x}_{0}$ and Gaussian noise, $\bm{\epsilon} \sim \mathcal{N}(\mathbf{0},\,\mathbf{I})$.
On the other hand, the denoising (reverse) step for an arbitrary step $t$ is written as:
\begin{equation}
    p_{\theta}(\mathbf{x}_{t-1}|\mathbf{x}_{t}) = \mathcal{N}(\mathbf{x}_{t-1};\bm{\mu}_{\theta}(\mathbf{x}_{t}, t), \mathbf{\Sigma}_{\theta}(\mathbf{x}_{t}, t))
\end{equation}
where $\mathbf{\Sigma}_{\theta}(\mathbf{x}_{t}, t)=\sigma_{t}^{2}\mathbf{I}$ is untrained time dependent constants in the original DDPM paper. With the epsilon-parametrization, the term of mean $\bm{\mu}_{\theta}(\mathbf{x}_{t}, t)$ is defined as follows:
\begin{equation}
    \bm{\mu}_{\theta}(\mathbf{x}_{t}, t) = \frac{1}{\sqrt{\alpha_{t}}} \left( \mathbf{x}_{t} - \frac{\beta_{t}}{\sqrt{1-\Bar{\alpha}_{t}}} \bm{\epsilon}_{\theta}(\mathbf{x}_{t}, t) \right)
\end{equation}
Here, $\bm{\epsilon}_{\theta}$ is a function that predicts the additive Guassian noise $\bm{\epsilon}$ from $\mathbf{x}_{t}$. Generally, in the diffusion model implementation, $\bm{\epsilon}_{\theta}$ can be defined as a network and is trained with the objective function as follows:
\begin{equation}
    \lVert \bm{\epsilon}
    - \bm{\epsilon}_{\theta} \left( \sqrt{\Bar{\alpha}_{t}} \mathbf{x}_{0} + \sqrt{1 - \Bar{\alpha}_{t}}\bm{\epsilon}, t \right) \rVert^{2}
    \label{equation:original_ddpm_loss}
\end{equation}
For arbitrary step $t$, therefore, we can sample $\mathbf{x}_{t-1}$ from $p_{\theta}(\mathbf{x}_{t-1}|\mathbf{x}_{t})$ using the noise estimation network $\bm{\epsilon}_{\theta}$.

\subsection{Wiener Filter-Based Model}
Diffusion models commonly aim to denoise the Gaussian noise from given $\mathbf{x}_{T}$ to generate a new image sample that belongs to the data distribution. 
If we assume that the denoising (reverse) process in the diffusion model is a linear operation, we can define the linear denoising Wiener filter $\mathbf{h}_{t}$ that optimizes $j_{t}=\lVert \sqrt{\Bar{\alpha}_t}\mathbf{x}_{0}-\mathbf{h}_{t}\ast \mathbf{x}_{t} \rVert^{2}$, where $\sqrt{\Bar{\alpha}_t}$ is a scaling factor. To determine the optimal Wiener filter, we manipulate $\mathbf{j}_{t}$ in the frequency domain.
\begin{align}
    J_{t}=\;&\lVert \sqrt{\Bar{\alpha}_t} \mathbf{X}_{0}-\mathbf{H}_{t} \cdot \mathbf{X}_{t} \rVert^{2} \\ 
    =\;&\lVert \sqrt{\Bar{\alpha}_t} \mathbf{X}_{0}-\mathbf{H}_{t} \cdot ( \sqrt{\Bar{\alpha}_t}\mathbf{X}_{0} + \sqrt{1 - \Bar{\alpha}_t} \bm{\mathcal{E}} ) \rVert^{2}
    \label{equation:wiener_filter_objective}
\end{align}
where $\mathbf{X}_{t}, \mathbf{H}_{t}$, and $\bm{\mathcal{E}}$ represent the frequency response of $\mathbf{x}_{t}, \mathbf{h}_{t}$, and $\bm{\epsilon}$, respectively. Note that $\mathbf{X}_{0}$ and $\bm{\mathcal{E}}$ are independent each other. To determine the optimal solution, we calculate the partial derivative of Eq.~\ref{equation:wiener_filter_objective} as follows:
\begin{align}
    \nonumber
    \frac{\partial J_{t}}{\partial\mathbf{H}_{t}}=\;& -2 \Bar{\alpha}_{t} \mathbf{X}_{0}^{\ast}\mathbf{X}_{0} + 2 \Bar{\alpha}_{t} \mathbf{X}_{0}^{\ast} \mathbf{H}_{t}^{\ast} \mathbf{X}_{0} + 2 (1-\Bar{\alpha}_{t}) \bm{\mathcal{E}}^{\ast} \mathbf{H}_{t}^{\ast} \bm{\mathcal{E}} \\
    =\;& 0
    \label{equation:partial_derivative}
\end{align}
From Eq.~\ref{equation:partial_derivative}, we can define the conjugate transposed Wiener filter in the frequency domain as follows:
\begin{align}
    \mathbf{H}_{t}^{\ast}=\frac{\Bar{\alpha}_{t} \lvert \mathbf{X}_{0} \rvert^2 } {\Bar{\alpha}_{t} \lvert \mathbf{X}_{0} \rvert^2 + (1-\Bar{\alpha}_{t}) \lvert \bm{\mathcal{E}} \rvert^2}
    \label{equation:optimal_filter}
\end{align}
Since $\lvert \mathbf{X}_{0} \rvert^2$ in Eq.~\ref{equation:optimal_filter} follows the power law \cite{powerlaw}, $\lvert \mathbf{X}_{0} \rvert^2 \approx 1/f^2$, and $\bm{\mathcal{E}}^{2}=1$ due to its definition. Then, the optimal Wiener filter can be rewritten as:
\begin{align}
    \mathbf{H}_{t}^{\ast}(f)=\frac{\Bar{\alpha}_{t}}{\Bar{\alpha}_{t} + (1-\Bar{\alpha}_{t}) \cdot f^2}
\end{align}

Now, we can re-express Eq.~\ref{equation:original_ddpm_loss}, an objective function of the diffusion model, using our Wiener filter-based analysis. From Eq.~\ref{equation:forward_process}, additive noise at an arbitrary timestep $t$ can be defined as:
\begin{equation}
    \bm{\epsilon} = \frac{\mathbf{x}_{t}-\sqrt{\Bar{\alpha}_{t}}\mathbf{x}_{0}}{\sqrt{1-\Bar{\alpha}_{t}}}
\end{equation}
Likewise, in terms of denoising of the reverse process using the Wiener filter which satisfies $j_{t} = 0$, the noise estimated from the network can be written as:
\begin{equation}
    \bm{\epsilon}_{\theta} = \frac{\mathbf{x}_{t}-\mathbf{h}_{t} \ast \mathbf{x}_{t}}{\sqrt{1-\Bar{\alpha}_{t}}}
\end{equation}
Finally, the Wiener filter-based objective function of the diffusion model is defined as:
\begin{equation}
    \lVert \bm{\epsilon}
    - \gamma \cdot (\mathbf{1}-\mathbf{h}_{t}) \ast \mathbf{x}_{t} \rVert^{2}
    \label{equation:wiener_filter_ddpm_loss}
\end{equation}
where $\gamma = 1/\sqrt{1-\Bar{\alpha}_{t}}$.

\section{Dataset Preparation}

\subsubsection{CycleGAN \cite{cyclegan}}
We take CycleGAN model from the officially released repository. We selected six pairs of datasets from their repository, \ie apple2orange, horse2zebra, summer2winter, and each case of the inverse. We trained the models with the setting of $256 \times 256$ resolution. Then, we take real and fake (translated) images from each dataset and model. Therefore, we prepared 8,269 real images and 8,269 fake images.

\subsubsection{StarGAN \cite{stargan}}
We take StarGAN model from the officially released repository. We trained the model on CelebA dataset \cite{celeba}. We resized images to $256 \times 256$ during the training. After the training step, we sampled 20,000 real images from the dataset and synthesized them using StarGAN model. Therefore, we prepared 20,000 real images and 20,000 fake images.

\subsubsection{StarGAN2 \cite{starganv2}}
We take StarGAN2 model from the officially released repository. We trained the model on AFHQ dataset \cite{starganv2}. We resized images to $256 \times 256$ during the training. After training, we generated fake images for the test dataset of AFHQ using StarGAN2 model. Therefore, we prepared 15,714 real images and 15,714 fake images.

\subsubsection{StyleGAN \cite{stylegan}}
We take StyleGAN model from the Pytorch-implemented repository. We trained the model on FFHQ dataset \cite{stylegan} with the setting of $256 \times 256$ resolution. We generated 20,000 fake images from the trained StyleGAN model. Then, we sampled 20,000 images from FFHQ dataset. Therefore, we prepared 20,000 real images and 20,000 fake images.

\subsubsection{DDPM \cite{ddpm}}
We take DDPM model from the officially released repository. We trained DDPM model on the two benchmarks, \ie FFHQ and LSUN Church \cite{stylegan, lsun}. During the training, we resized images to $128 \times 128$. From the real images, we sampled 16,000 images to build the dataset. And then, we generated 16,000 images from the DDPM model. Therefore, we prepared 16,000 real images and 16,000 fake images for the DDPM model on FFHQ and the model on LSUN Church, respectively.

\begin{figure*}[!t]
\centering
    \begin{tabular}{C{0.42\columnwidth}C{0.42\columnwidth}C{0.42\columnwidth}C{0.42\columnwidth}}
    \includegraphics[width=0.45\columnwidth]{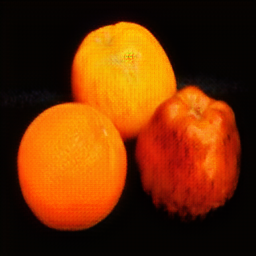} &
    \includegraphics[width=0.45\columnwidth]{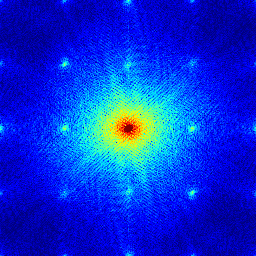} &
    \includegraphics[width=0.45\columnwidth]{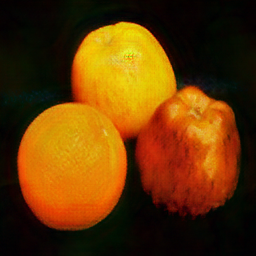} &
    \includegraphics[width=0.45\columnwidth]{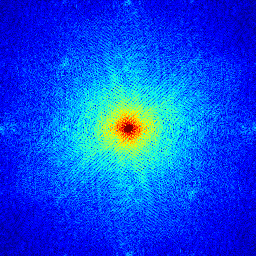} \\
    \includegraphics[width=0.45\columnwidth]{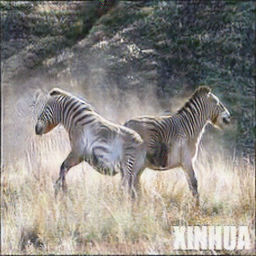} &
    \includegraphics[width=0.45\columnwidth]{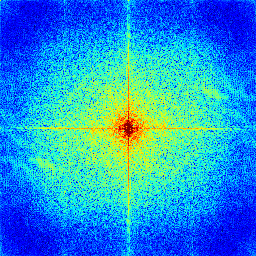} &
    \includegraphics[width=0.45\columnwidth]{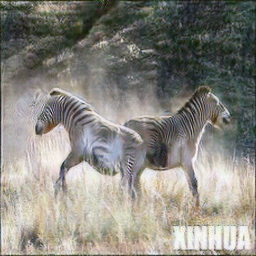} &
    \includegraphics[width=0.45\columnwidth]{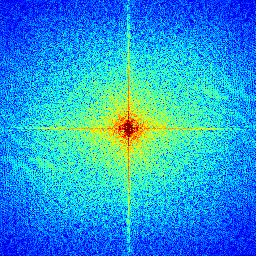} \\
    \includegraphics[width=0.45\columnwidth]{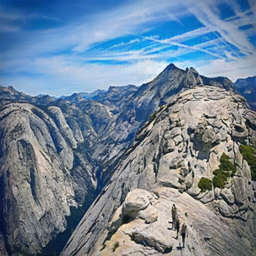} &
    \includegraphics[width=0.45\columnwidth]{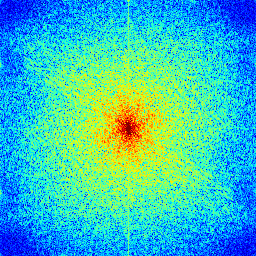} &
    \includegraphics[width=0.45\columnwidth]{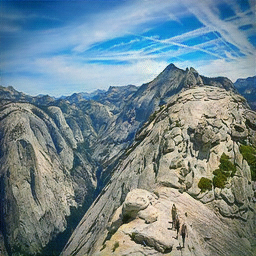} &
    \includegraphics[width=0.45\columnwidth]{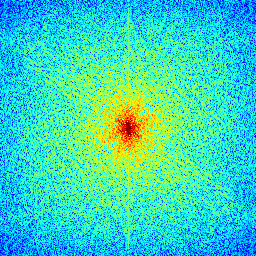} \\
    \includegraphics[width=0.45\columnwidth]{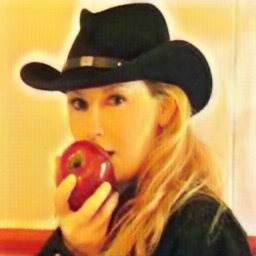} &
    \includegraphics[width=0.45\columnwidth]{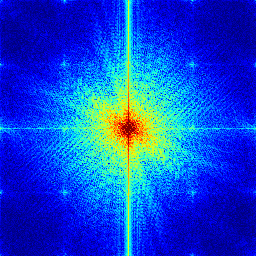} &
    \includegraphics[width=0.45\columnwidth]{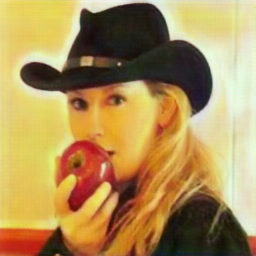} &
    \includegraphics[width=0.45\columnwidth]{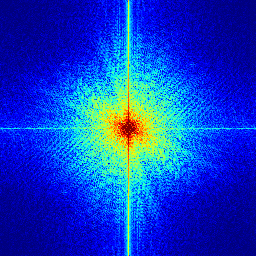} \\
    \includegraphics[width=0.45\columnwidth]{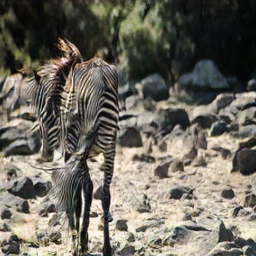} &
    \includegraphics[width=0.45\columnwidth]{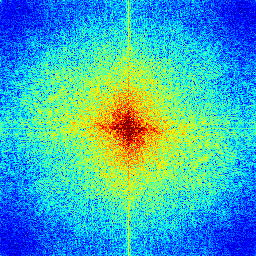} &
    \includegraphics[width=0.45\columnwidth]{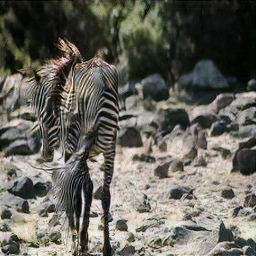} &
    \includegraphics[width=0.45\columnwidth]{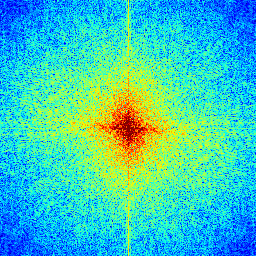} \\
    \multicolumn{2}{c}{Original Generated} & \multicolumn{2}{c}{STIG}
    \end{tabular}
    \caption{Original generated and STIG-refined Image-Spectrum pairs of CycleGAN \cite{cyclegan} benchmark.}
    \label{figure:supp_cyclegan}
\end{figure*}

\subsubsection{DDIM \cite{ddim}}
We employ the pre-trained model on FFHQ and LSUN Church benchmark in the case of DDPM. We used the sampled real images in DDPM real/fake dataset. Then, we generated 16,000 images using the DDIM sampling method with the pre-trained DDPM model. Therefore, we prepared 16,000 real images and 16,000 fake images for the DDIM model on FFHQ and the model on LSUN Church, respectively.

\section{Visual Examples}
In this section, we provide more visual examples of the proposed method. For benchmarks used in our paper, we showcase the results of STIG not only in the frequency domain but also in the image domain. Note that the warmer color implies a higher intensity in the frequency spectrum.

\begin{figure*}[!t]
\centering
    \begin{tabular}{C{0.42\columnwidth}C{0.42\columnwidth}C{0.42\columnwidth}C{0.42\columnwidth}}
    \includegraphics[width=0.45\columnwidth]{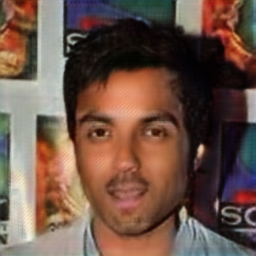} &
    \includegraphics[width=0.45\columnwidth]{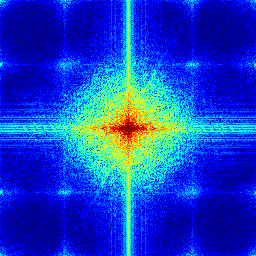} &
    \includegraphics[width=0.45\columnwidth]{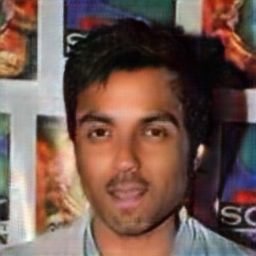} &
    \includegraphics[width=0.45\columnwidth]{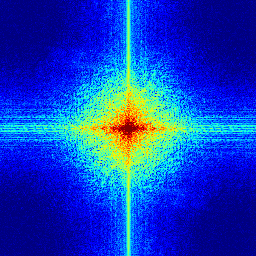} \\
    \includegraphics[width=0.45\columnwidth]{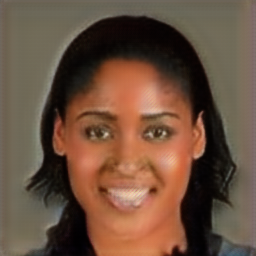} &
    \includegraphics[width=0.45\columnwidth]{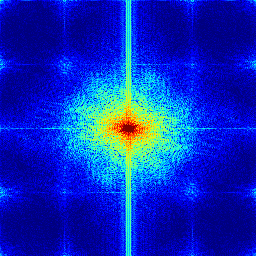} &
    \includegraphics[width=0.45\columnwidth]{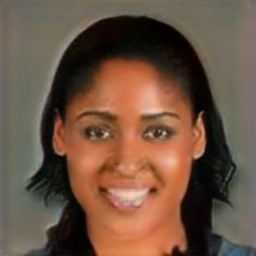} &
    \includegraphics[width=0.45\columnwidth]{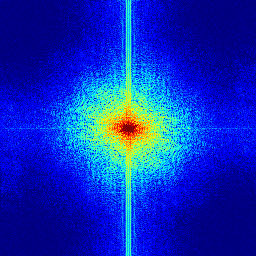} \\
    \includegraphics[width=0.45\columnwidth]{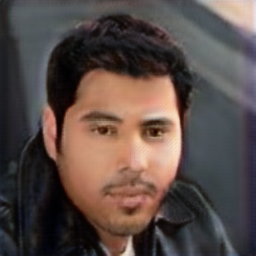} &
    \includegraphics[width=0.45\columnwidth]{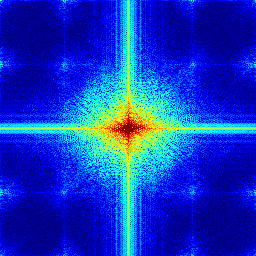} &
    \includegraphics[width=0.45\columnwidth]{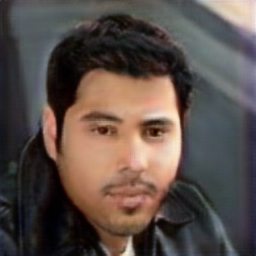} &
    \includegraphics[width=0.45\columnwidth]{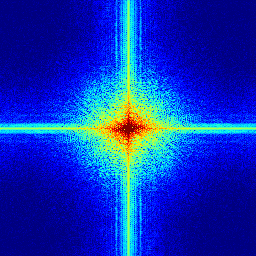} \\
    \includegraphics[width=0.45\columnwidth]{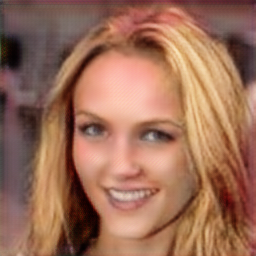} &
    \includegraphics[width=0.45\columnwidth]{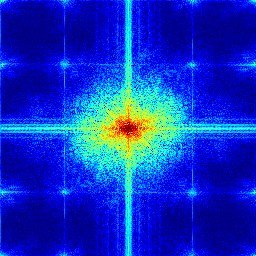} &
    \includegraphics[width=0.45\columnwidth]{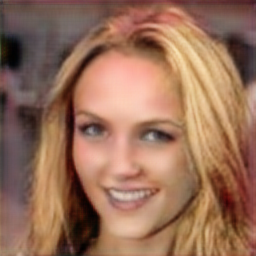} &
    \includegraphics[width=0.45\columnwidth]{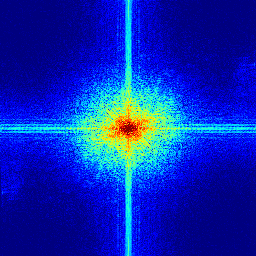} \\
    \includegraphics[width=0.45\columnwidth]{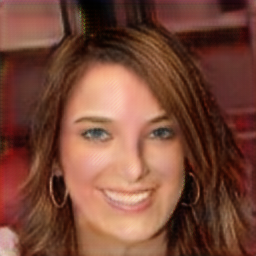} &
    \includegraphics[width=0.45\columnwidth]{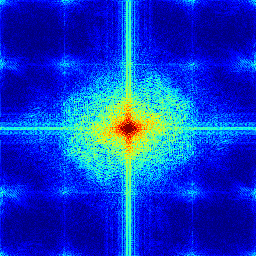} &
    \includegraphics[width=0.45\columnwidth]{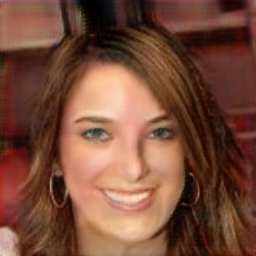} &
    \includegraphics[width=0.45\columnwidth]{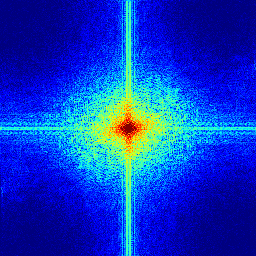} \\
    \multicolumn{2}{c}{Original Generated} & \multicolumn{2}{c}{STIG}
    \end{tabular}
    \caption{Original generated and STIG-refined Image-Spectrum pairs of StarGAN \cite{stargan} benchmark.}
    \label{figure:supp_stargan}
\end{figure*}
\clearpage

\begin{figure*}[!t]
\centering
    \begin{tabular}{C{0.42\columnwidth}C{0.42\columnwidth}C{0.42\columnwidth}C{0.42\columnwidth}}
    \includegraphics[width=0.45\columnwidth]{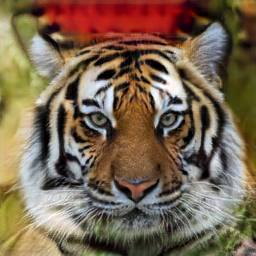} &
    \includegraphics[width=0.45\columnwidth]{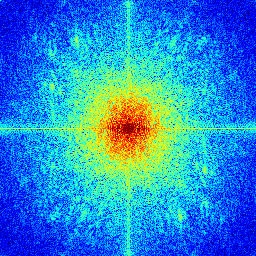} &
    \includegraphics[width=0.45\columnwidth]{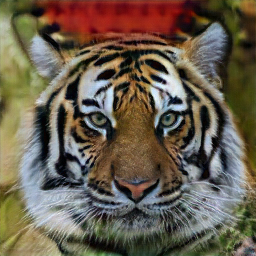} &
    \includegraphics[width=0.45\columnwidth]{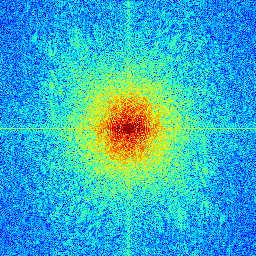} \\
    \includegraphics[width=0.45\columnwidth]{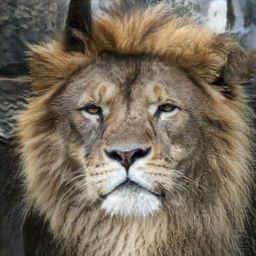} &
    \includegraphics[width=0.45\columnwidth]{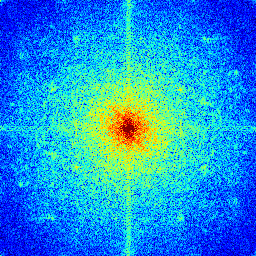} &
    \includegraphics[width=0.45\columnwidth]{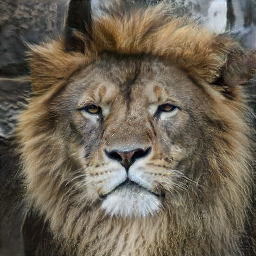} &
    \includegraphics[width=0.45\columnwidth]{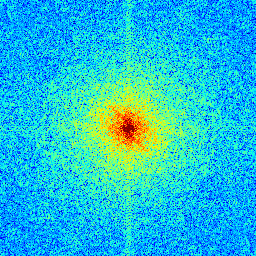} \\
    \includegraphics[width=0.45\columnwidth]{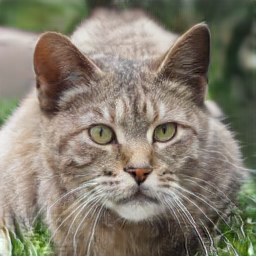} &
    \includegraphics[width=0.45\columnwidth]{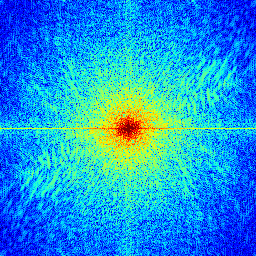} &
    \includegraphics[width=0.45\columnwidth]{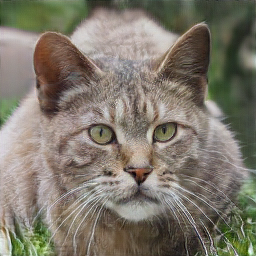} &
    \includegraphics[width=0.45\columnwidth]{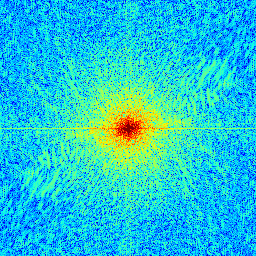} \\
    \includegraphics[width=0.45\columnwidth]{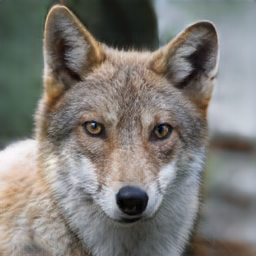} &
    \includegraphics[width=0.45\columnwidth]{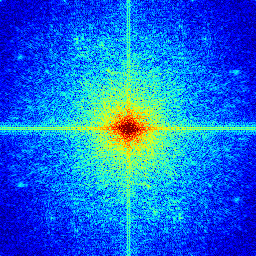} &
    \includegraphics[width=0.45\columnwidth]{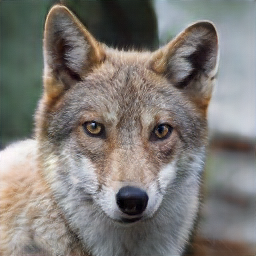} &
    \includegraphics[width=0.45\columnwidth]{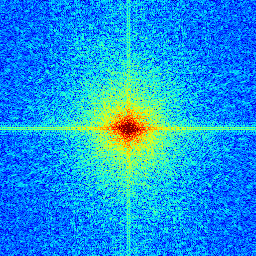} \\
    \includegraphics[width=0.45\columnwidth]{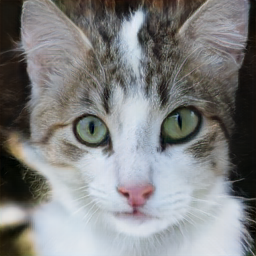} &
    \includegraphics[width=0.45\columnwidth]{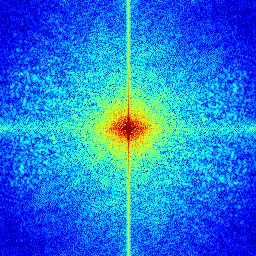} &
    \includegraphics[width=0.45\columnwidth]{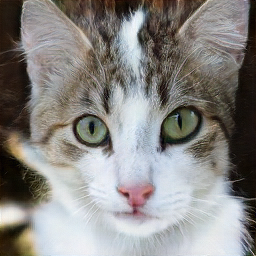} &
    \includegraphics[width=0.45\columnwidth]{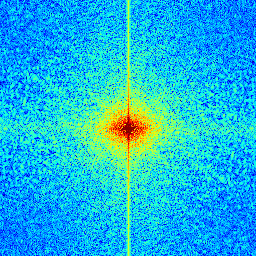} \\
    \multicolumn{2}{c}{Original Generated} & \multicolumn{2}{c}{STIG}
    \end{tabular}
    \caption{Original generated and STIG-refined Image-Spectrum pairs of StarGAN2 \cite{starganv2} benchmark.}
    \label{figure:supp_stargan2}
\end{figure*}
\clearpage

\begin{figure*}[!t]
\centering
    \begin{tabular}{C{0.42\columnwidth}C{0.42\columnwidth}C{0.42\columnwidth}C{0.42\columnwidth}}
    \includegraphics[width=0.45\columnwidth]{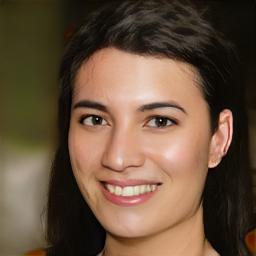} &
    \includegraphics[width=0.45\columnwidth]{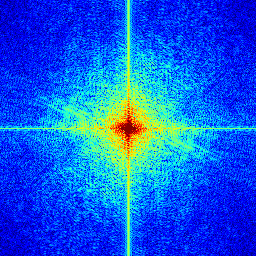} &
    \includegraphics[width=0.45\columnwidth]{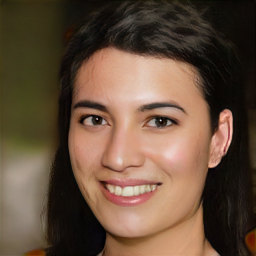} &
    \includegraphics[width=0.45\columnwidth]{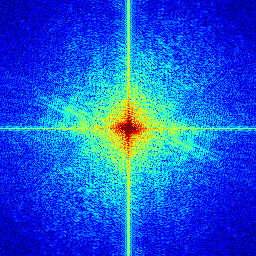} \\
    \includegraphics[width=0.45\columnwidth]{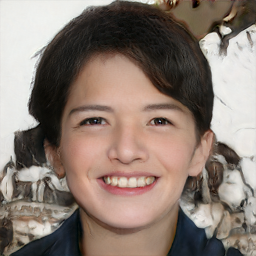} &
    \includegraphics[width=0.45\columnwidth]{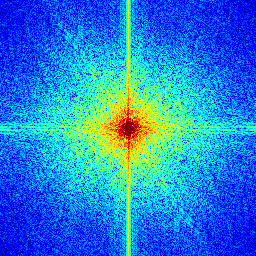} &
    \includegraphics[width=0.45\columnwidth]{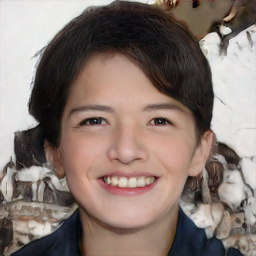} &
    \includegraphics[width=0.45\columnwidth]{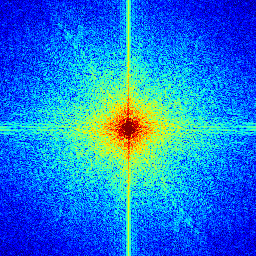} \\
    \includegraphics[width=0.45\columnwidth]{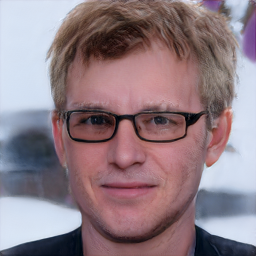} &
    \includegraphics[width=0.45\columnwidth]{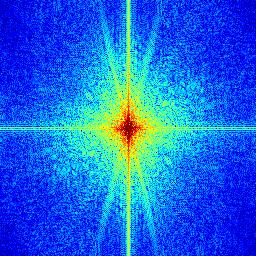} &
    \includegraphics[width=0.45\columnwidth]{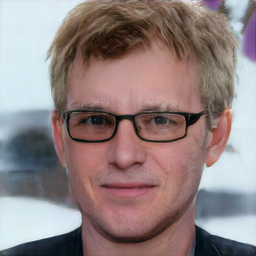} &
    \includegraphics[width=0.45\columnwidth]{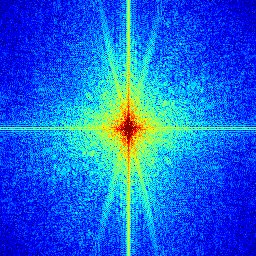} \\
    \includegraphics[width=0.45\columnwidth]{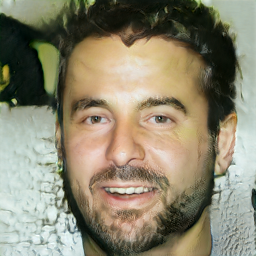} &
    \includegraphics[width=0.45\columnwidth]{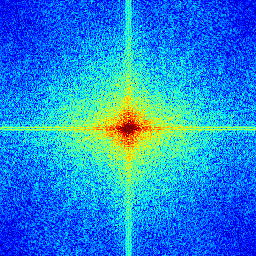} &
    \includegraphics[width=0.45\columnwidth]{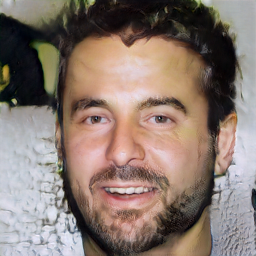} &
    \includegraphics[width=0.45\columnwidth]{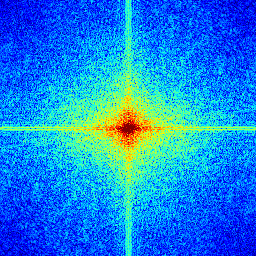} \\
    \includegraphics[width=0.45\columnwidth]{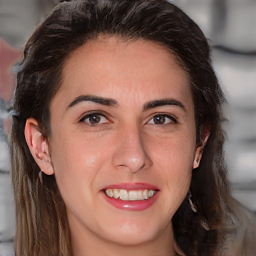} &
    \includegraphics[width=0.45\columnwidth]{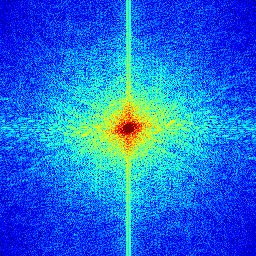} &
    \includegraphics[width=0.45\columnwidth]{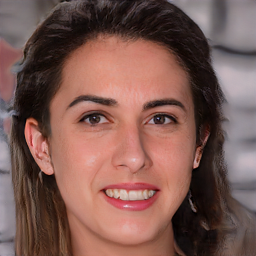} &
    \includegraphics[width=0.45\columnwidth]{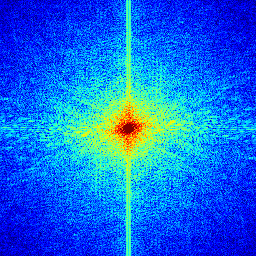} \\
    \multicolumn{2}{c}{Original Generated} & \multicolumn{2}{c}{STIG}
    \end{tabular}
    \caption{Original generated and STIG-refined Image-Spectrum pairs of StyleGAN \cite{stylegan} benchmark.}
    \label{figure:supp_stylegan}
\end{figure*}
\clearpage

\begin{figure*}[!t]
\centering
    \begin{tabular}{C{0.42\columnwidth}C{0.42\columnwidth}C{0.42\columnwidth}C{0.42\columnwidth}}
    \includegraphics[width=0.45\columnwidth]{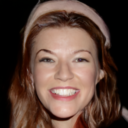} &
    \includegraphics[width=0.45\columnwidth]{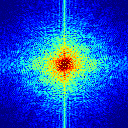} &
    \includegraphics[width=0.45\columnwidth]{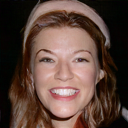} &
    \includegraphics[width=0.45\columnwidth]{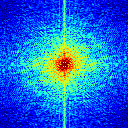} \\
    \includegraphics[width=0.45\columnwidth]{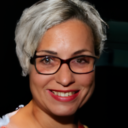} &
    \includegraphics[width=0.45\columnwidth]{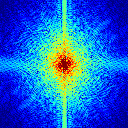} &
    \includegraphics[width=0.45\columnwidth]{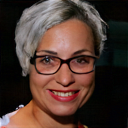} &
    \includegraphics[width=0.45\columnwidth]{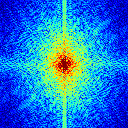} \\
    \includegraphics[width=0.45\columnwidth]{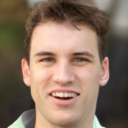} &
    \includegraphics[width=0.45\columnwidth]{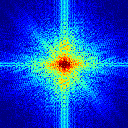} &
    \includegraphics[width=0.45\columnwidth]{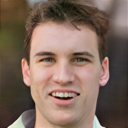} &
    \includegraphics[width=0.45\columnwidth]{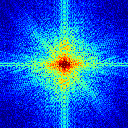} \\
    \includegraphics[width=0.45\columnwidth]{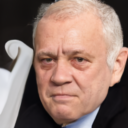} &
    \includegraphics[width=0.45\columnwidth]{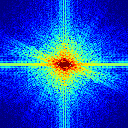} &
    \includegraphics[width=0.45\columnwidth]{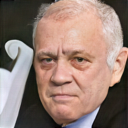} &
    \includegraphics[width=0.45\columnwidth]{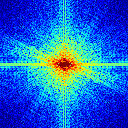} \\
    \includegraphics[width=0.45\columnwidth]{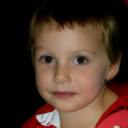} &
    \includegraphics[width=0.45\columnwidth]{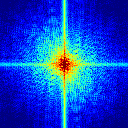} &
    \includegraphics[width=0.45\columnwidth]{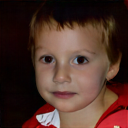} &
    \includegraphics[width=0.45\columnwidth]{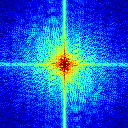} \\
    \multicolumn{2}{c}{Original Generated} & \multicolumn{2}{c}{STIG}
    \end{tabular}
    \caption{Original generated and STIG-refined Image-Spectrum pairs of DDPM-Face \cite{ddpm} benchmark.}
    \label{figure:supp_ddpm_face}
\end{figure*}
\clearpage

\begin{figure*}[!t]
\centering
    \begin{tabular}{C{0.42\columnwidth}C{0.42\columnwidth}C{0.42\columnwidth}C{0.42\columnwidth}}
    \includegraphics[width=0.45\columnwidth]{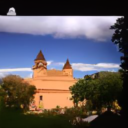} &
    \includegraphics[width=0.45\columnwidth]{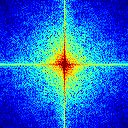} &
    \includegraphics[width=0.45\columnwidth]{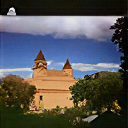} &
    \includegraphics[width=0.45\columnwidth]{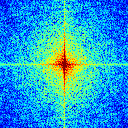} \\
    \includegraphics[width=0.45\columnwidth]{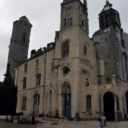} &
    \includegraphics[width=0.45\columnwidth]{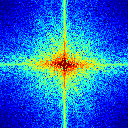} &
    \includegraphics[width=0.45\columnwidth]{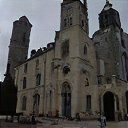} &
    \includegraphics[width=0.45\columnwidth]{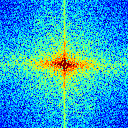} \\
    \includegraphics[width=0.45\columnwidth]{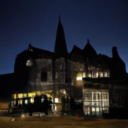} &
    \includegraphics[width=0.45\columnwidth]{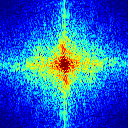} &
    \includegraphics[width=0.45\columnwidth]{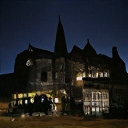} &
    \includegraphics[width=0.45\columnwidth]{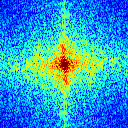} \\
    \includegraphics[width=0.45\columnwidth]{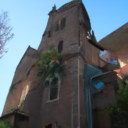} &
    \includegraphics[width=0.45\columnwidth]{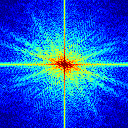} &
    \includegraphics[width=0.45\columnwidth]{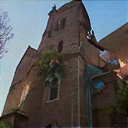} &
    \includegraphics[width=0.45\columnwidth]{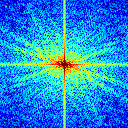} \\
    \includegraphics[width=0.45\columnwidth]{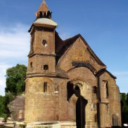} &
    \includegraphics[width=0.45\columnwidth]{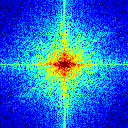} &
    \includegraphics[width=0.45\columnwidth]{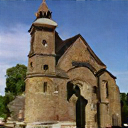} &
    \includegraphics[width=0.45\columnwidth]{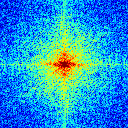} \\
    \multicolumn{2}{c}{Original Generated} & \multicolumn{2}{c}{STIG}
    \end{tabular}
    \caption{Original generated and STIG-refined Image-Spectrum pairs of DDPM-Church \cite{ddpm} benchmark.}
    \label{figure:supp_ddpm_church}
\end{figure*}
\clearpage

\begin{figure*}[!t]
\centering
    \begin{tabular}{C{0.42\columnwidth}C{0.42\columnwidth}C{0.42\columnwidth}C{0.42\columnwidth}}
    \includegraphics[width=0.45\columnwidth]{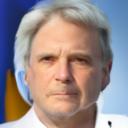} &
    \includegraphics[width=0.45\columnwidth]{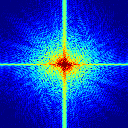} &
    \includegraphics[width=0.45\columnwidth]{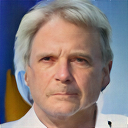} &
    \includegraphics[width=0.45\columnwidth]{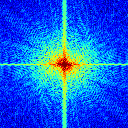} \\
    \includegraphics[width=0.45\columnwidth]{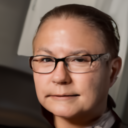} &
    \includegraphics[width=0.45\columnwidth]{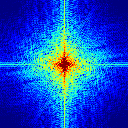} &
    \includegraphics[width=0.45\columnwidth]{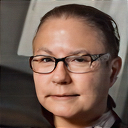} &
    \includegraphics[width=0.45\columnwidth]{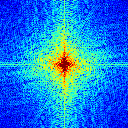} \\
    \includegraphics[width=0.45\columnwidth]{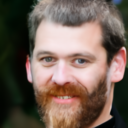} &
    \includegraphics[width=0.45\columnwidth]{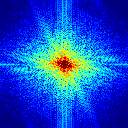} &
    \includegraphics[width=0.45\columnwidth]{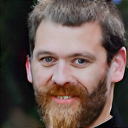} &
    \includegraphics[width=0.45\columnwidth]{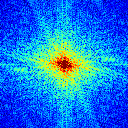} \\
    \includegraphics[width=0.45\columnwidth]{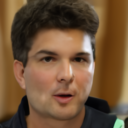} &
    \includegraphics[width=0.45\columnwidth]{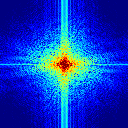} &
    \includegraphics[width=0.45\columnwidth]{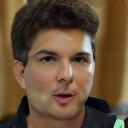} &
    \includegraphics[width=0.45\columnwidth]{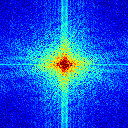} \\
    \includegraphics[width=0.45\columnwidth]{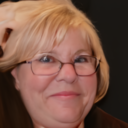} &
    \includegraphics[width=0.45\columnwidth]{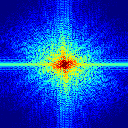} &
    \includegraphics[width=0.45\columnwidth]{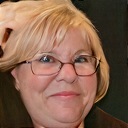} &
    \includegraphics[width=0.45\columnwidth]{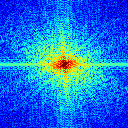} \\
    \multicolumn{2}{c}{Original Generated} & \multicolumn{2}{c}{STIG}
    \end{tabular}
    \caption{Original generated and STIG-refined Image-Spectrum pairs of DDIM-Face \cite{ddim} benchmark.}
    \label{figure:supp_ddim_face}
\end{figure*}
\clearpage

\begin{figure*}[!t]
\centering
    \begin{tabular}{C{0.42\columnwidth}C{0.42\columnwidth}C{0.42\columnwidth}C{0.42\columnwidth}}
    \includegraphics[width=0.45\columnwidth]{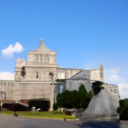} &
    \includegraphics[width=0.45\columnwidth]{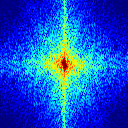} &
    \includegraphics[width=0.45\columnwidth]{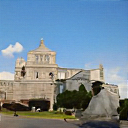} &
    \includegraphics[width=0.45\columnwidth]{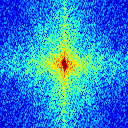} \\
    \includegraphics[width=0.45\columnwidth]{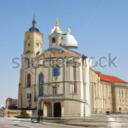} &
    \includegraphics[width=0.45\columnwidth]{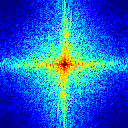} &
    \includegraphics[width=0.45\columnwidth]{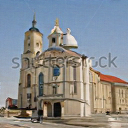} &
    \includegraphics[width=0.45\columnwidth]{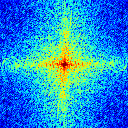} \\
    \includegraphics[width=0.45\columnwidth]{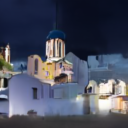} &
    \includegraphics[width=0.45\columnwidth]{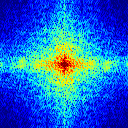} &
    \includegraphics[width=0.45\columnwidth]{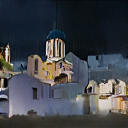} &
    \includegraphics[width=0.45\columnwidth]{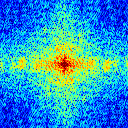} \\
    \includegraphics[width=0.45\columnwidth]{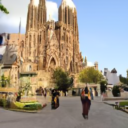} &
    \includegraphics[width=0.45\columnwidth]{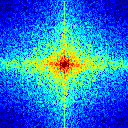} &
    \includegraphics[width=0.45\columnwidth]{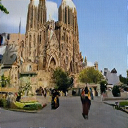} &
    \includegraphics[width=0.45\columnwidth]{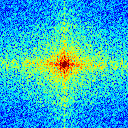} \\
    \includegraphics[width=0.45\columnwidth]{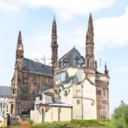} &
    \includegraphics[width=0.45\columnwidth]{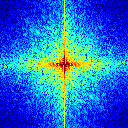} &
    \includegraphics[width=0.45\columnwidth]{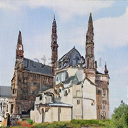} &
    \includegraphics[width=0.45\columnwidth]{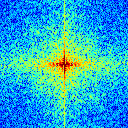} \\
    \multicolumn{2}{c}{Original Generated} & \multicolumn{2}{c}{STIG}
    \end{tabular}
    \caption{Original generated and STIG-refined Image-Spectrum pairs of DDIM-Church \cite{ddim} benchmark.}
    \label{figure:supp_ddim_church}
\end{figure*}
\clearpage

\begin{figure*}[!t]
\centering
    \begin{tabular}{@{\hskip -0.3cm}C{0.01\columnwidth}C{0.24\columnwidth}C{0.24\columnwidth}C{0.24\columnwidth}C{0.24\columnwidth}C{0.24\columnwidth}C{0.24\columnwidth}C{0.24\columnwidth}}
    \rotatebox{90}{CycleGAN} &
    \includegraphics[width=0.27\columnwidth]{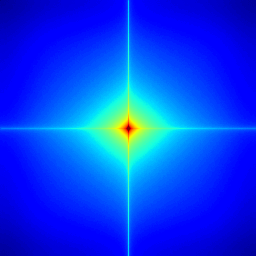} &
    \includegraphics[width=0.27\columnwidth]{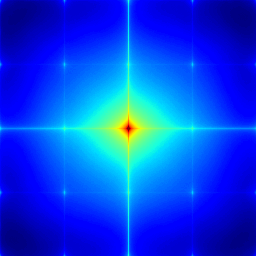} &
    \includegraphics[width=0.27\columnwidth]{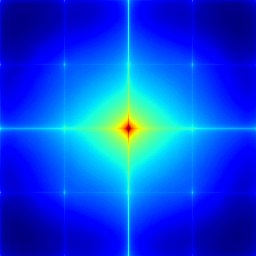} &
    \includegraphics[width=0.27\columnwidth]{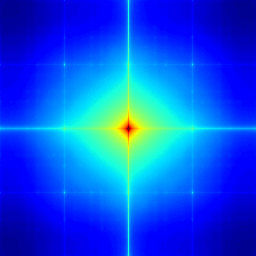} &
    \includegraphics[width=0.27\columnwidth]{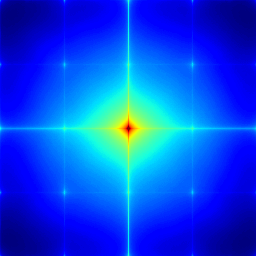} &
    \includegraphics[width=0.27\columnwidth]{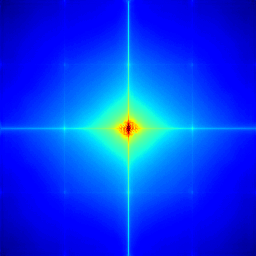} &
    \includegraphics[width=0.27\columnwidth]{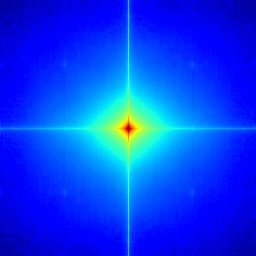} \\

    \rotatebox{90}{StarGAN} &
    \includegraphics[width=0.27\columnwidth]{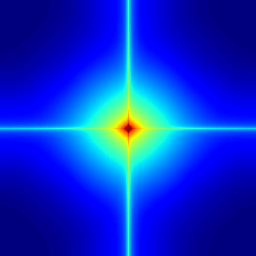} &
    \includegraphics[width=0.27\columnwidth]{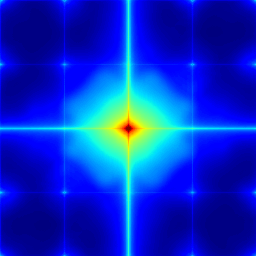} &
    \includegraphics[width=0.27\columnwidth]{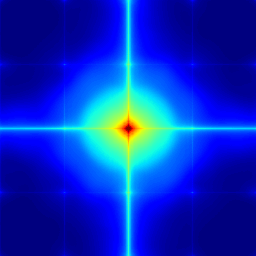} &
    \includegraphics[width=0.27\columnwidth]{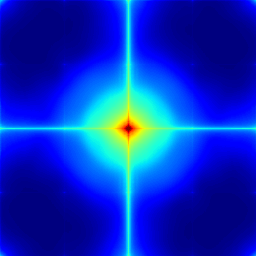} &
    \includegraphics[width=0.27\columnwidth]{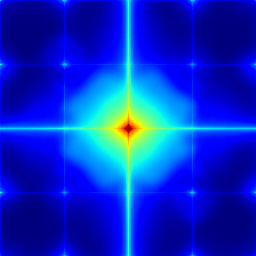} &
    \includegraphics[width=0.27\columnwidth]{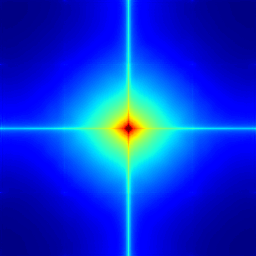} &
    \includegraphics[width=0.27\columnwidth]{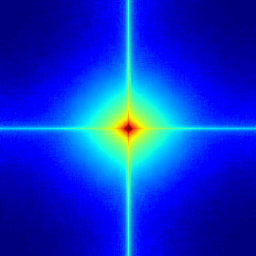} \\

    \rotatebox{90}{StarGAN2} &
    \includegraphics[width=0.27\columnwidth]{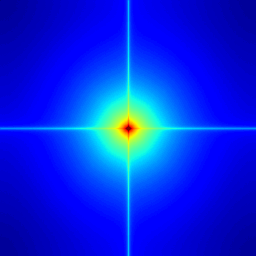} &
    \includegraphics[width=0.27\columnwidth]{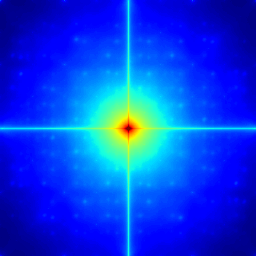} &
    \includegraphics[width=0.27\columnwidth]{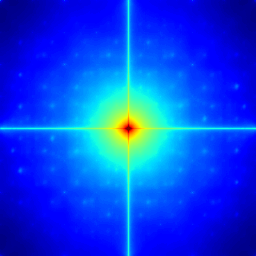} &
    \includegraphics[width=0.27\columnwidth]{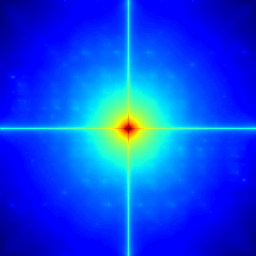} &
    \includegraphics[width=0.27\columnwidth]{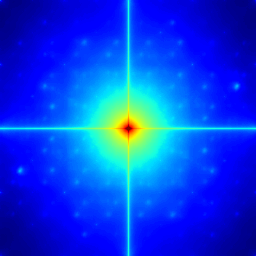} &
    \includegraphics[width=0.27\columnwidth]{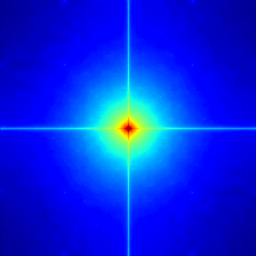} &
    \includegraphics[width=0.27\columnwidth]{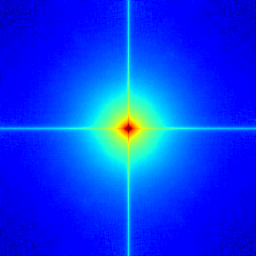} \\

    \rotatebox{90}{StyleGAN} &
    \includegraphics[width=0.27\columnwidth]{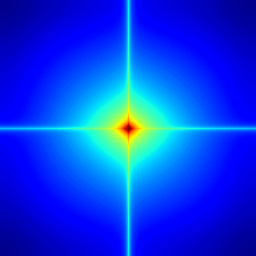} &
    \includegraphics[width=0.27\columnwidth]{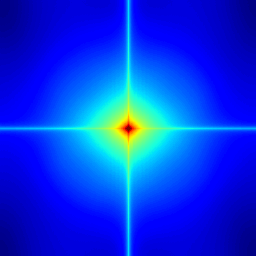} &
    \includegraphics[width=0.27\columnwidth]{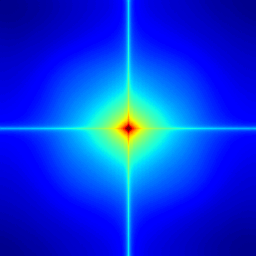} &
    \includegraphics[width=0.27\columnwidth]{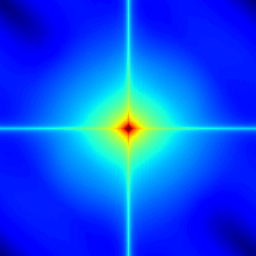} &
    \includegraphics[width=0.27\columnwidth]{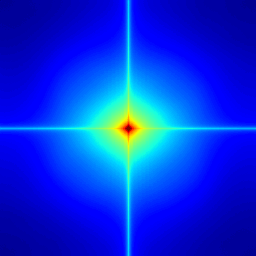} &
    \includegraphics[width=0.27\columnwidth]{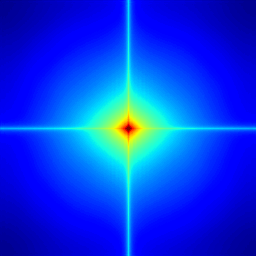} &
    \includegraphics[width=0.27\columnwidth]{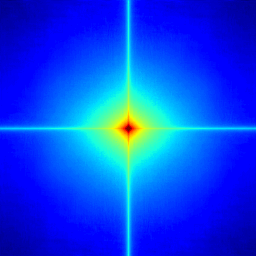} \\
    {} & {\;\;Real} & {\;Original Generated} & {\;\;Durall \etal} & {\;Jung and Keuper} & {\;\;FFL} & {\;\,SpectralGAN} & {\;\;STIG}
    \end{tabular}
    \caption{Averaged magnitude of the spectrum for generative adversarial networks.}
    \label{figure:average_spectrum}
\end{figure*}

\begin{figure*}[!t]
\centering
    \begin{tabular}{@{\hskip -0.3cm}C{0.01\columnwidth}C{0.24\columnwidth}C{0.24\columnwidth}C{0.24\columnwidth}}
    \rotatebox{90}{DDPM-Face} &
    \includegraphics[width=0.27\columnwidth]{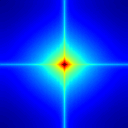} &
    \includegraphics[width=0.27\columnwidth]{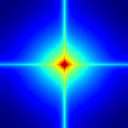} &
    \includegraphics[width=0.27\columnwidth]{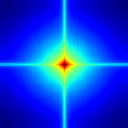} \\

    \rotatebox{90}{DDPM-Church} &
    \includegraphics[width=0.27\columnwidth]{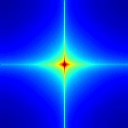} &
    \includegraphics[width=0.27\columnwidth]{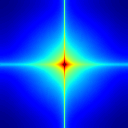} &
    \includegraphics[width=0.27\columnwidth]{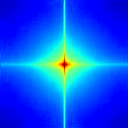} \\

    \rotatebox{90}{DDIM-Face} &
    \includegraphics[width=0.27\columnwidth]{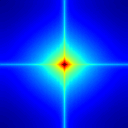} &
    \includegraphics[width=0.27\columnwidth]{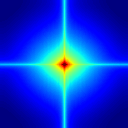} &
    \includegraphics[width=0.27\columnwidth]{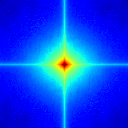} \\

    \rotatebox{90}{DDIM-Church} &
    \includegraphics[width=0.27\columnwidth]{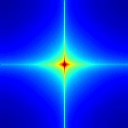} &
    \includegraphics[width=0.27\columnwidth]{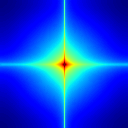} &
    \includegraphics[width=0.27\columnwidth]{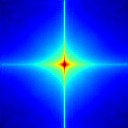} \\
    {} & {\;\;Real} & {\;Original Generated} & {\;\;STIG}
    \end{tabular}
    \caption{Averaged magnitude of the spectrum for diffusion models.}
    \label{figure:average_spectrum}
\end{figure*}
\clearpage

\bibliography{aaai24}

\end{document}